\documentclass[sn-mathphys-num]{sn-jnl}
 
\usepackage{graphicx}%
\usepackage{multirow}%
\usepackage{amsmath,amssymb,amsfonts}%
\usepackage{amsthm}%
\usepackage{mathrsfs}%
\usepackage[title]{appendix}%
\usepackage{xcolor}%
\usepackage{textcomp}%
\usepackage{manyfoot}%
\usepackage{booktabs}%
\usepackage{algorithm}%
\usepackage{algorithmicx}%
\usepackage{algpseudocode}%
\usepackage{listings}%

\theoremstyle{thmstyleone}%
\theoremstyle{thmstyletwo}%

\theoremstyle{thmstylethree}%

\usepackage{subfig}
\usepackage{bm}

\usepackage{amsmath}
\usepackage{amssymb}

\usepackage{booktabs}

\begin{document}

\title[Lester: rotoscope animation through video object segmentation and tracking]{Lester: rotoscope animation through video object segmentation and tracking}


\author*[1]{\fnm{Ruben} \sur{Tous}}\email{ruben.tous@upc.edu}

\affil*[1]{\orgdiv{Department of Computer Architecture}, \orgname{Universitat Politècnica de Catalunya (UPC)}, \orgaddress{\street{Jordi Girona, 1-3}, \city{Barcelona}, \postcode{08034}, \country{Spain}}}


\abstract{This article introduces Lester, a novel method to automatically synthetise retro-style 2D animations from videos. The method approaches the challenge mainly as an object segmentation and tracking problem. Video frames are processed with the Segment Anything Model (SAM) and the resulting masks are tracked through subsequent frames with DeAOT, a method of hierarchical propagation for semi-supervised video object segmentation. The geometry of the masks' contours is simplified with the Douglas-Peucker algorithm. Finally, facial traits, pixelation and a basic shadow effect can be optionally added. The results show that the method exhibits an excellent temporal consistency and can correctly process videos with different poses and appearances, dynamic shots, partial shots and diverse backgrounds. The proposed method provides a more simple and deterministic approach than diffusion models based video-to-video translation pipelines, which suffer from temporal consistency problems and do not cope well with pixelated and schematic outputs. The method is also much most practical than techniques based on 3D human pose estimation, which require custom handcrafted 3D models and are very limited with respect to the type of scenes they can process.}




\maketitle

\section{Introduction}

%
%
In industries such as cinema or video games, rotoscoping is employed to leverage the information from live-action footage for the creation of 2D animations. This technique contributed to the emergence of a unique visual aesthetic in certain iconic video games from the 80s and 90s, such as ``Prince of Persia'' (1989), ``Another World'' (1991) or ``Flashback'' (1992), and it continues to be employed in modern titles like ``Lunark'' (2023). However, as in the case of Lunark, this technique still requires manual work at the frame level, making it very expensive. This article describes a novel method, called Lester, to automate this process. A user enters a video and a color palette and the system automatically generates a 2D retro-style animation in a format directly usable in a video game or in an animated movie. 

\begin{figure}
\centering
\subfloat{\includegraphics[width=1.65in]{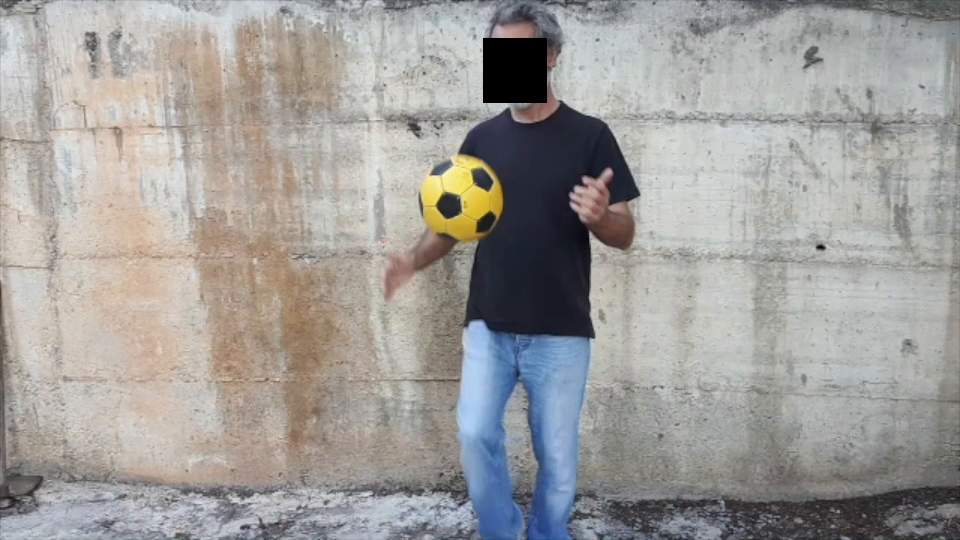} \label{fig:ex1}}
\subfloat{\includegraphics[width=1.65in]{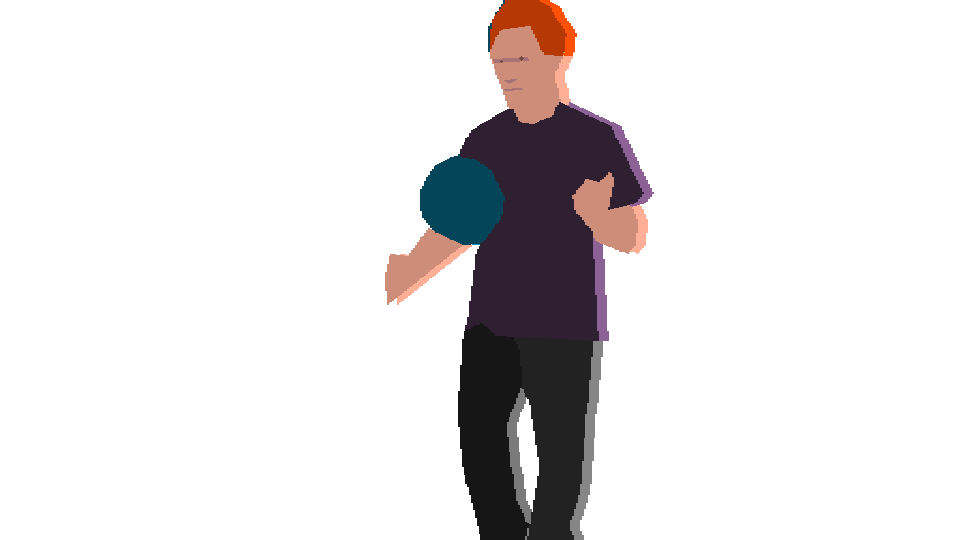} \label{fig:ex2}}\\
\caption{Example synthetised animation frame (right) and its related original frame (left, face has been anonymized) from an example input video.}
\end{figure}

The method approaches the challenge mainly as an object segmentation problem. Given a frame depicting a person, a mask for each different visual trait (hair, skin, clothes, etc.) is generated with the Segment Anything Model (SAM) \cite{sam}. To univocally identify the masks in all the frames, a video object tracking algorithm is employed using the approach of (\cite{sam-track}). The masks are purged to solve segmentation errors and their geometry is simplified with the Douglas-Peucker algorithm \cite{douglas-peucker} to achieve the desired visual style. Finally, facial traits, pixelation and a basic shadow effect can be optionally added.

This being essentially a specific case of unsupervised video-to-video translation, a different strategy could involve employing conditional generative models, like diffusion models. Yet, as we will explore subsequently, the proposed method proves to be a superior alternative due to its enhanced efficacy in handling the specific visual style and addressing temporal consistency issues. 

\begin{figure}
\centering
\includegraphics[width=3.75in]{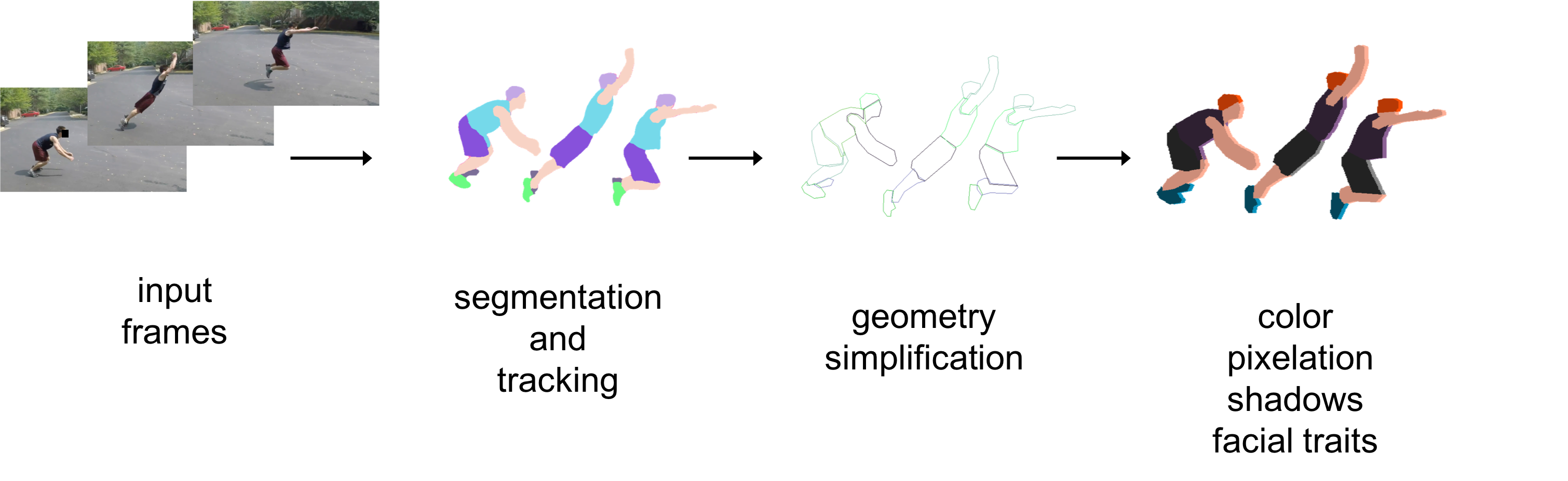} \\
\caption{Overview of the system's workflow (inputs on the left, output on the right)}
\label{fig:workflow}
\end{figure}

\section{Related Work}

Max Fleischer's invention of the rotoscope in 1915 transformed the ability to create animated cartoons by using still or moving images. Variants of this technique have been used to this day in animated films (e.g. ``A Scanner Darkly'', (2006)),  television animated series (e.g. ``Undone'' (2019)) and video games (e.g. ``Prince of Persia'' (1989), ``Another World'' (1991), ``Flashback'' (1992), ``Lunark'' (2023)). Contemporary productions integrate image processing algorithms to facilitate the task, but they are mainly based on costly manual work. Achieving full automation with a polished professional outcome (uniform style, spatial and temporal consistency) remains unfeasible at present. However, recent advances in Artificial Intelligence (AI) opens the possibility to attempt advancing to fully automation. 

To the best of our knowledge, there are two main strategies to address the problem. On the one hand, the challenge can be addressed with conditional generative models, prevalent in image translation tasks nowadays. The first methods (e.g. CartoonGAN \cite{cartoongan}, AnimeGAN\cite{animegan} or the work in \cite{cartoonizationLiu2022}) used conditional generative adversarial networks (GAN) to transform photos of real-world scenes into cartoon style images. These methods have recently been surpassed by those based on diffusion models such as Stable Diffusion\cite{stablediffusion}. Although most of these methods have been designed for text-to-image generation, they have been successfully adapted for image-to-image translation using techniques such as SDEdit \cite{image2image} or ControlNet \cite{controlnet}. Pure image-to-image translation methods with diffusion models also exists (e.g. SR3 \cite{sr3} and Palette \cite{palette}), but with a supervised setup that can only be applied to situations in which a groundtruth of pairs of images (original, target) such as super-resolution or colorization. In any case, even if we have a good conditional generative model, here is still the challenge of applying it to the frames of a video while preserving temporal consistency. Latest attempts (e.g. \cite{yang2023rerender}) combine tricks (e.g. tiled inputs) to uniformize the output of the generative model with video-guided style interpolation techniques (e.g. EbSynth \cite{ebsynth}). The results of these methods have improved greatly, and for some styles the results are so good that they are already used in independent productions. However, for most digital art styles these methods impose an undesirable trade-off between style fidelity and temporal consistency. 

On the other hand, a completely different approach consists on applying a 3D human pose estimation algorithm to track the 3D pose through the different frames. The 3D pose is then applied to a 3D character model with the desired aspect (hair, clothes, etc.). Typically this 3D character model is created by hand with a specialised software. Finally, a proper shader and rendering configuration are setup to obtain the desired visual style. This is the approach taken in \cite{video2cartoon1}, \cite{video2cartoon2}, \cite{photowakeup}, \cite{pictonaut} and \cite{completepose}. This approach has many drawbacks. First, it depends on the state-of-art of 3D human pose estimation, which has advanced a lot but still fails in many situations (low resolution, truncations, uncommon poses) mainly due to the difficulty to obtain enough groundtruth data. Second, it requires to design a proper 3D model for each different character. Third, the 3D pose is based on a given model, typically SMPL. To apply it to a different model it requires a pose retargetting step. The joint rotations can be easily retargetted but the SMPL blend shapes not, resulting in unrealistic body deformations. This is why demos of this kind of works are usually done with robots, as the don't have deformable parts. An advantage of this approach with respect to generative models is that it provides a good temporal consistency, as inconsistencies from the 3D human pose estimation step can be properly solved with interpolation such as is done in \cite{pictonaut}. 

In this work we propose a third approach, mainly based on segmentation and tracking. Unlike the first mentioned strategy (generative models), the proposed approach is more predictable and provides much better temporal consistency. Unlike the second strategy mentioned (3D human pose estimation), the proposed approach is more robust and much more practical since it does not require having a custom 3D model for each new character. This is not the first time that this kind of problem has been addressed this way. In \cite{Fiser17-SIG} segmentation was used to stylize a portrait video given an artistic example image obtaining better results than neural style transfer \cite{styletransfer}, prevalent at that time. 

\section{Methodology}

\subsection{Overview}

The main input to the method is a target video sequence $T$ of a human performance. There are not general constraints about the traits of the depicted person (gender, age, clothes, etc.), pose, orientation, camera movements, background, lighting, occlusions, truncations (partial body shots) or resolution. However, certain scenarios are known to potentially impact the results negatively (this will be addressed in Section ~\ref{results}). While the method is primarily designed for videos featuring a single person, its theoretical application with multiple individuals will be discussed in Section ~\ref{results}. When operating without additional inputs, the method defaults to specific settings for segmentation, colors, and other attributes. However, it is anticipated that the user might prefer specifying custom configurations, necessitating supplementary inputs such as segmentation prompts, color palettes, etc., as detailed in corresponding subsections below. The output of the method is a set of PNG images showing the resulting animated character and no background (where the alpha channel is set to zero).

The method consists of three independent processing steps. The main one is the segmentation and tracking of the different visual traits (hair, skin, clothes, etc.) of the person featured in the video frames. The second step focuses on rectifying segmentation errors through mask purging and simplifying their polygonal contours. The third and final step involves applying customizable finishing details, including colors, faux shadows, facial features, and pixelation. Figure ~\ref{fig:workflow} shows the overall workflow of the method. Each component is elaborated upon in the subsequent subsections.

\subsection{Segmentation and tracking}

In order to segment the different visual traits of the character the Segment Anything Model (SAM) \cite{sam} is used. SAM is a promptable segmentation model which is becoming a foundation model to solve a variety of tasks. Given an input image with three channels (height x width x 3) and a prompt, that can include spatial or text information, it generates a mask (height x weight x 3), with each pixel assigned an RGB color. Each color represents an object label from the object label space $L = l_1, l_2, ...$, inferred from the prompt,  corresponding to the predicted object type at the corresponding pixel location in the input image. The prompt can be any information indicating what to segment in an image (e.g. a set of foreground/background points, a rough box or mask or free-form text). SAM consists on a image encoder (based on  MAE \cite{MAE} a pre-trained Vision Transformer \cite{ViT}) that computes an image embedding, a prompt encoder (being the free-form text based on CLIP \cite{CLIP}) that embeds prompts, and then the two information sources are combined in a mask decoder (based on Transformer segmentation models \cite{segmentation_transformers}) that predicts segmentation masks.

Applying SAM to all the frames of the input video is not enough as each time SAM employs different object labels. To do it consistently we use SAM-Track \cite{sam-track}. SAM results for the first frame are propagated to the second frame, with the same labels, with DeAOT \cite{DeAOT}, a method of hierarchical propagation for semi-supervised video object segmentation. The hierarchical propagation can gradually propagate information from past frames to the current frame and transfer the current frame feature from object-agnostic to object-specific. SAM is applied to subsequent frames to detect new objects, but the segmentation of already seen objects is carried out by DeAOT. 

\begin{figure}
\centering
\includegraphics[width=4.25in]{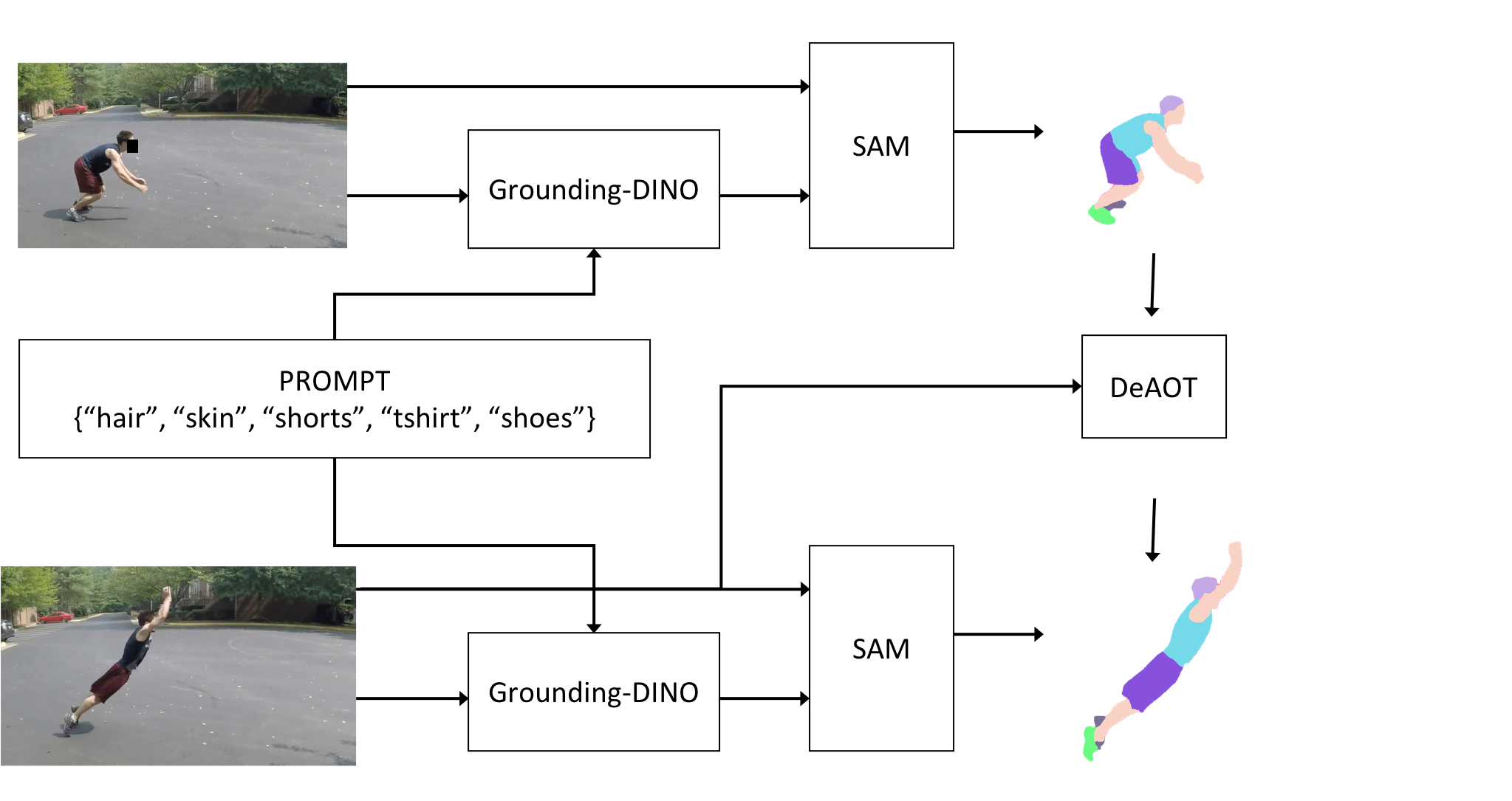} \\
\caption{Workflow of the segmentation and tracking stage for the first two frames (inputs on the left, output on the right).}
\label{fig:workflow-samtrack} 
\end{figure}

We have chosen keyword-based prompts to specify the desired segmentation. We provide some default prompts for common scenarios (e.g. {'hair', 'skin', 'shoes', 't-shirt', 'trousers'}) but it's anticipated that in many scenarios, the user will need to specify an alternative. These prompts are not processed with the free-text capabilities of SAM because of their limitations, instead they are processed with Grounding-DINO \cite{Grounding-DINO}, an open-set object detector, which translate them into object boxes that are sent to SAM as box prompts. One problem we have identified is that certain object combinations lead to incorrect maks when they are specified in the same prompt. To solve this, we allow specifying multiple prompts that are processed independently. Figure ~\ref{fig:workflow-samtrack} shows the workflow of the segmentation and tracking stage.

\subsection{Contours simplification}

\begin{figure}
\centering 
\subfloat[]{\includegraphics[width=0.6in]{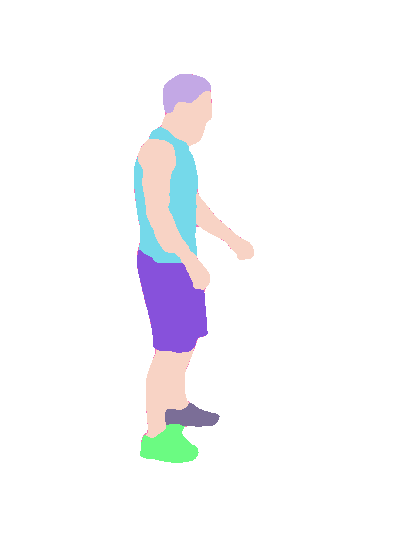} \label{fig:ex1}}
\subfloat[]{\includegraphics[width=0.6in]{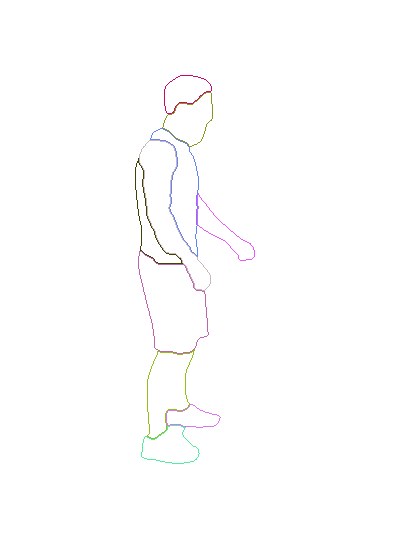} \label{fig:ex1}}
\subfloat[]{\includegraphics[width=0.6in]{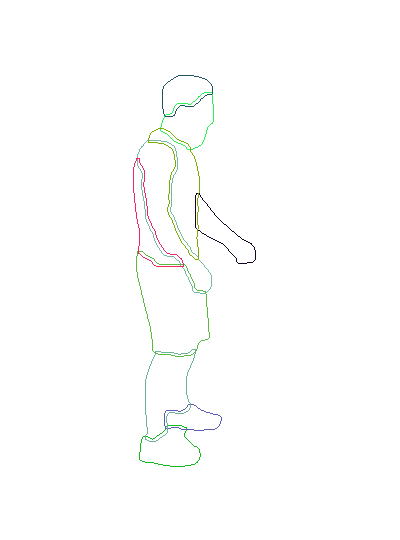} \label{fig:ex2}}
\subfloat[]{\includegraphics[width=0.6in]{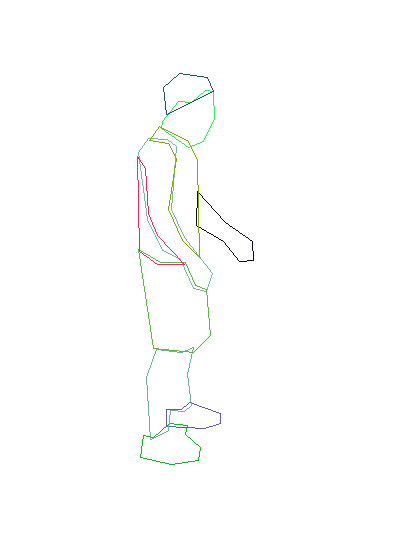} \label{fig:ex3}} 
\subfloat[]{\includegraphics[width=0.6in]{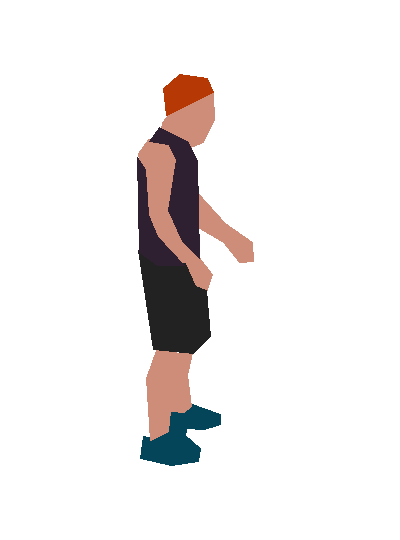} \label{fig:ex3}} 
\caption{Graphical overview of the contour simplification stage of one frame: (a) Result of the segmentation and tracking step (b) Contours detection (c) Contours dilation (d) Contours simplification (e) Resulting colored frame.} 
\label{fig:simplification} 
\end{figure}

One of the desired choices for the target visual style is the possibility to display very simple geometries, i.e. shape contours with short polygonal chains, in order to mimic the visual aspect of, e.g., ``Another World''. To achieve it, a series of steps are carried out. For each frame, the result of the segmentation and tracking step is a mask with each pixel assigned an RGB color representing an object label. The mask is first divided into submasks, one for each color/label. Each submask is divided, in turn, into contours with the algorithm from \cite{findcontours}. Each individual contour is represented with a vector of the vertex coordinates of the boundary line segments. The number of line segments of each contour is reduced with the Douglas-Peucker algorithm \cite{douglas-peucker}. The algorithm starts by joining the endpoints of the contour with a straight line. Vertices closer to the line than a given tolerance value $t$ are removed. Next, the line is split at the point farthest from the reference line, creating two new lines. The algorithm works recursively until all vertices within the tolerance are removed. Tolerance $t$ is one of the optional parameters of the proposed method and let the user control the level of simplification. As each contour is simplified independently, undesirable spaces can appear between the geometries of bordering contours that previously were perfectly aligned. To avoid this, and before the simplification step, all contours are dilated convolving the submask with a 4x4 kernel. It is worth mentioning that he simplification algorithm may fail for certain contour geometries and values of $t$. In that situations (usually very small contours) the original contour is kept. See Algorithm~\ref{alg:simplification} and Figure ~\ref{fig:simplification}.



\begin{algorithm}[H] 
\caption{Contour simplification algorithm}
\label{alg:simplification}
\begin{algorithmic}[1]
\Require{$M$ a list containing submasks, $\alpha$ a minimum contour area, $t$ a tolerance value for simplification}
\Ensure{$R$ a list of contours where $R_{i,j}$ is the simplified contour $j$ of submask $i$}
\Statex
\Function{Simplify}{$M, \alpha, t$}
    \For{$i \gets 0$ to $length(M)-1$}  
			\State {$C_i \gets getContours(M_i)$}
			\For{$j \gets 0$ to $length(C_i)-1$} 
				\If {$areaOf(C_{i,j}) > \alpha$}
		        	\State {$D_{i,j} \gets dilateContour(C_{i,j})$}
		        	\If {$contourCanBeSimplified(D_{i,j}, t)$}
		        		\State {$R_{i,j} \gets simplifyContour(D_{i,j}, t)$}
					\Else
						\State {$R_{i,j} \gets D_{i,j}$} 
		        	\EndIf
		        \EndIf 
	        \EndFor
    \EndFor
    \State \Return $R$
\EndFunction
\end{algorithmic}
\end{algorithm}

\subsection{Finishing details}

\begin{figure}
\centering
\subfloat[]{\includegraphics[width=0.8in]{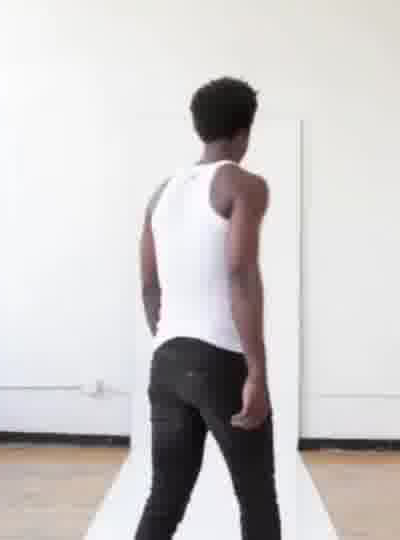} \label{fig:ex1}} 
\subfloat[]{\includegraphics[width=0.8in]{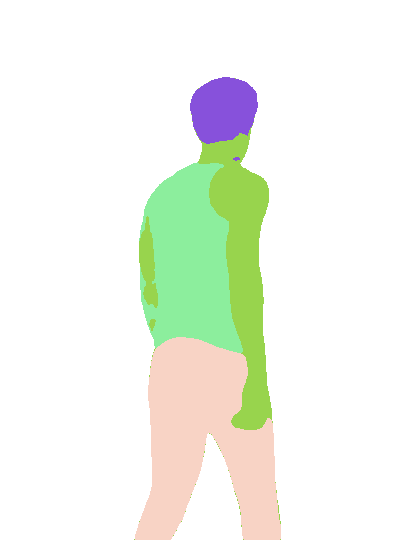} \label{fig:ex1}}
\subfloat[]{\includegraphics[width=0.8in]{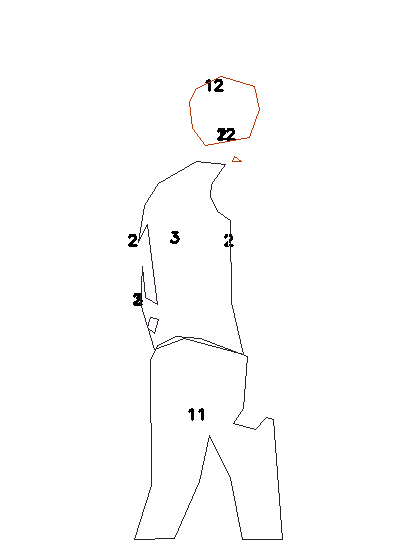} \label{fig:ex2}}
\subfloat[]{\includegraphics[width=0.8in]{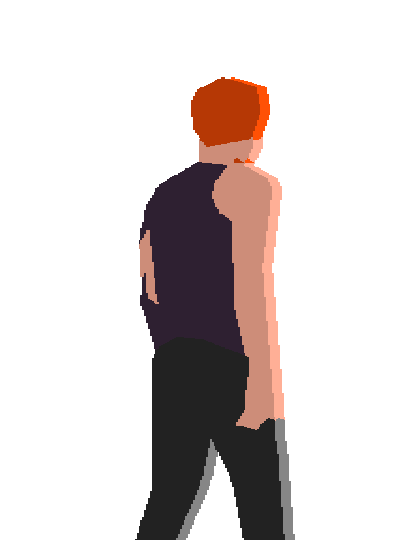} \label{fig:ex3}} 
\caption{Coloring process: (a) Input frame (b) Segmentation (c) Visual guide with unique identifiers for each color segment (d) Final frame.} 
\label{fig:color_ids} 
\end{figure}

The third and final step of the method entails applying customizable finishing details, granting users flexibility in shaping the visual style of the outcome. The most important aspect is the color. The method generates a visual guide with unique identifiers for each color segment, as shown in Figure ~\ref{fig:color_ids}. One of the inputs to the method is a color palette, which indicates the RGB values that will be applied to each segment identifier.

\begin{figure}
\centering
\subfloat[]{\includegraphics[width=0.7in]{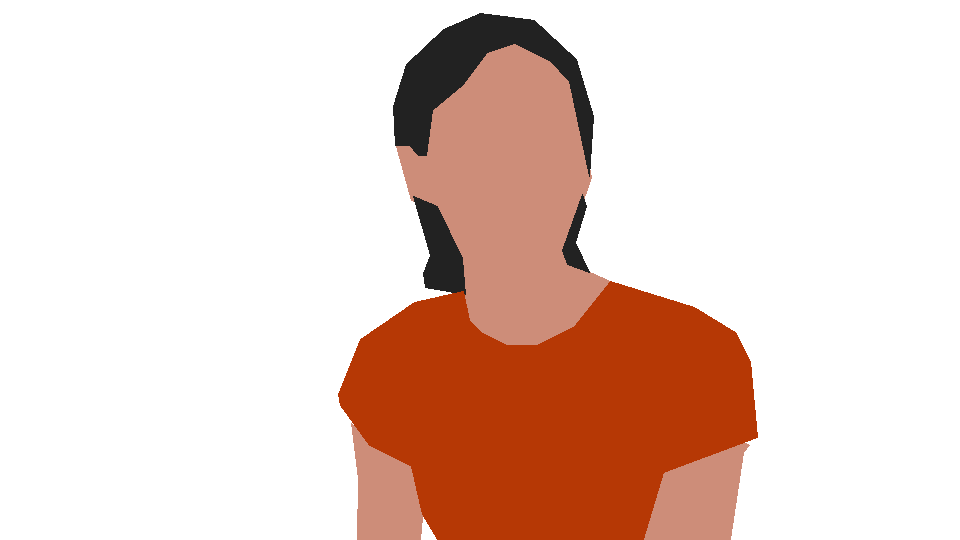} \label{fig:ex1}}
\subfloat[]{\includegraphics[width=0.7in]{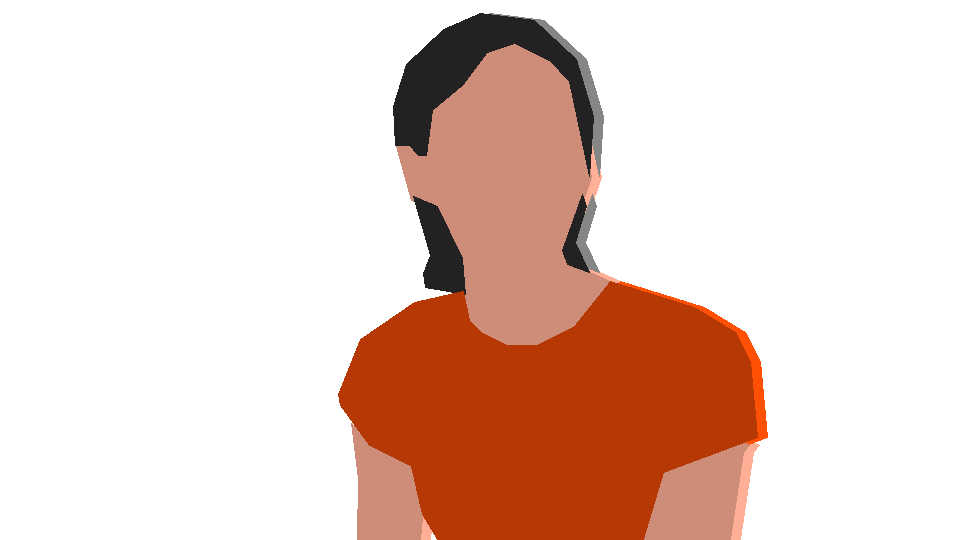} \label{fig:ex1}}
\subfloat[]{\includegraphics[width=0.7in]{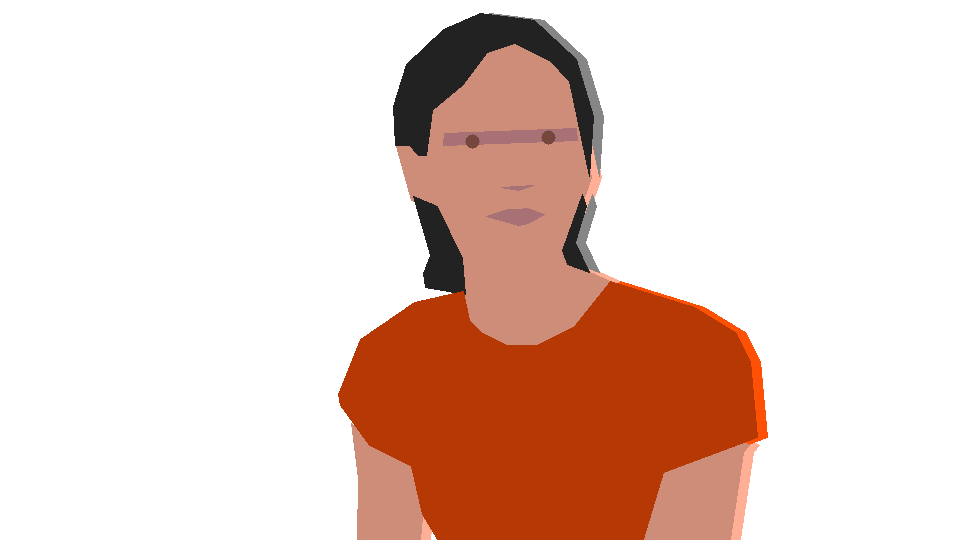} \label{fig:ex2}}
\subfloat[]{\includegraphics[width=0.7in]{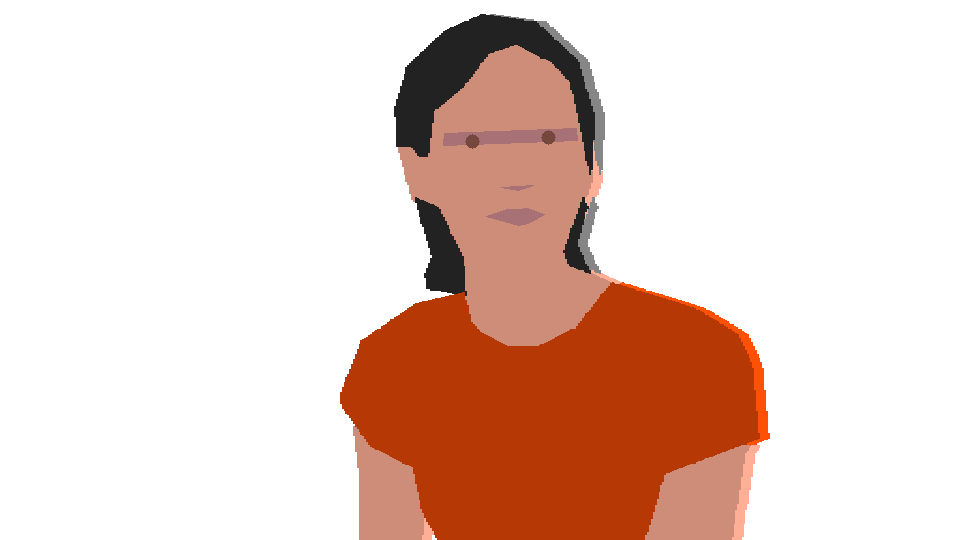} \label{fig:ex3}} 
\caption{Optional finishing details: (a) Flat method result (b) Shadows (c) Facial features (d) Pixelation.} 
\label{fig:finishing} 
\end{figure}

 Another finishing detail, which is optional, is the possibility of adding some schematic-looking facial features. The features are generated from the 68 facial landmarks that can be obtained using the frontal face detector described in \cite{dlib_HOG2}, which employs a Histogram of Oriented Gradients (HOG) \cite{HOG} combined with a linear classifier, an image pyramid, and a sliding window detection scheme. Another option is the possibility of adding a very simple shadow effect by superimposing the same image with different brightness and displaced horizontally. Finally, a pixelation effect is achieved by resizing the image down using bilinear interpolation and scaling it back up with nearest neighbour interpolation. Figure ~\ref{fig:finishing} shows how shadows, facial features and pixelation are progressively applied to an example frame.

\section{Experiments and results}
\label{results}

\begin{figure}
\centering
\subfloat{\includegraphics[width=0.60in]{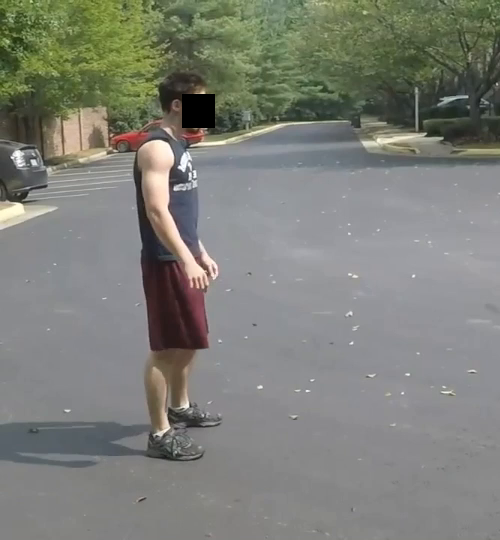} \label{fig:ex1}}
\subfloat{\includegraphics[width=0.60in]{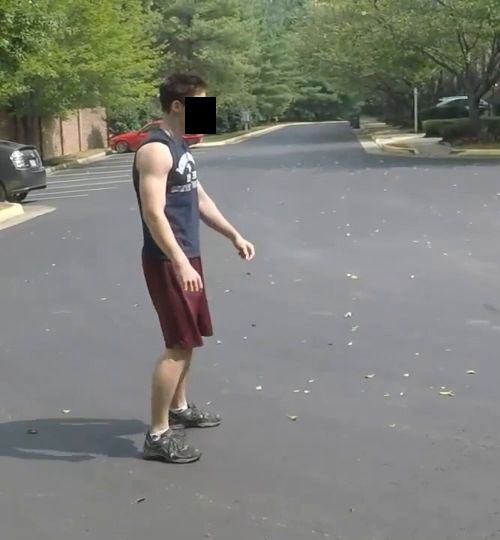} \label{fig:ex1}}
\subfloat{\includegraphics[width=0.60in]{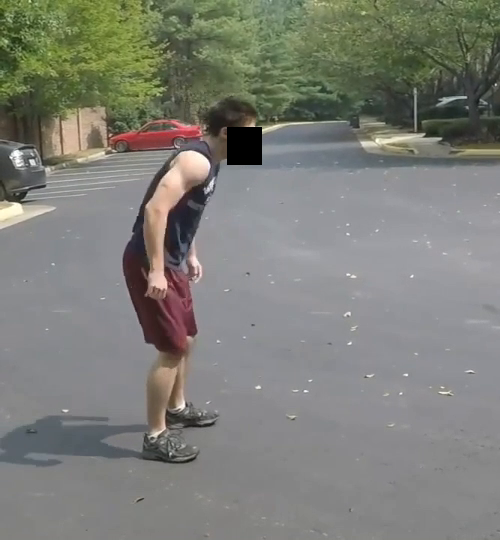} \label{fig:ex1}}
\subfloat{\includegraphics[width=0.60in]{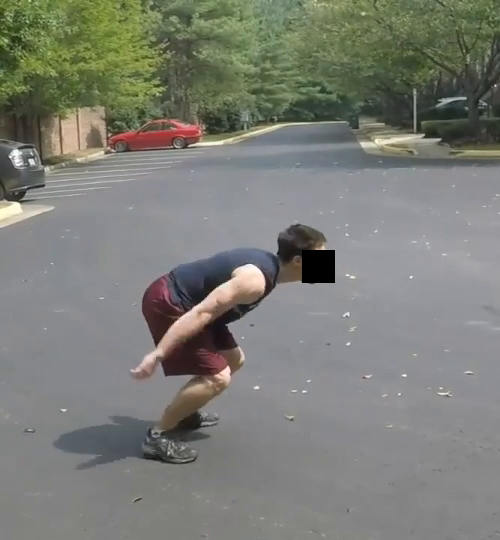} \label{fig:ex1}}
\subfloat{\includegraphics[width=0.60in]{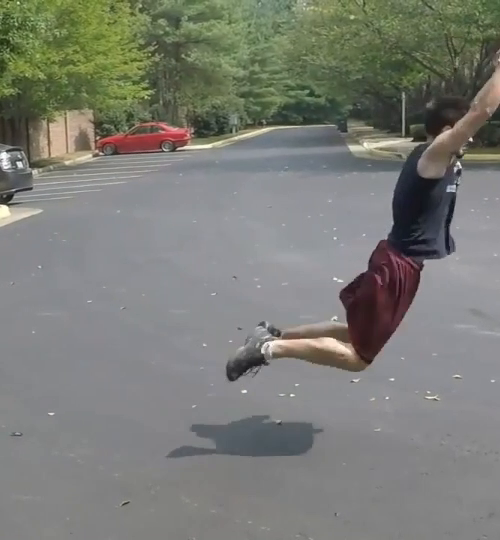} \label{fig:ex1}}\\

\subfloat{\includegraphics[width=0.60in]{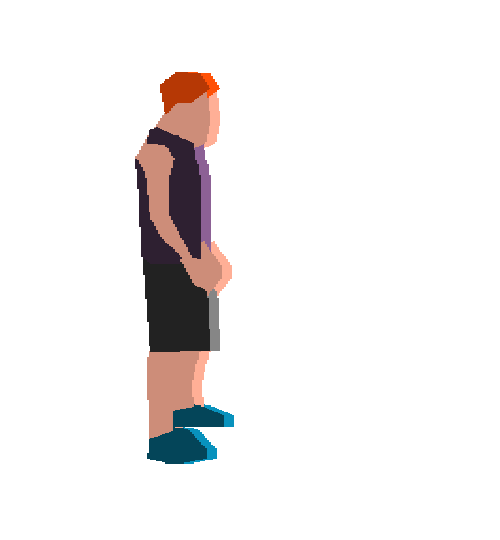} \label{fig:ex1}}
\subfloat{\includegraphics[width=0.60in]{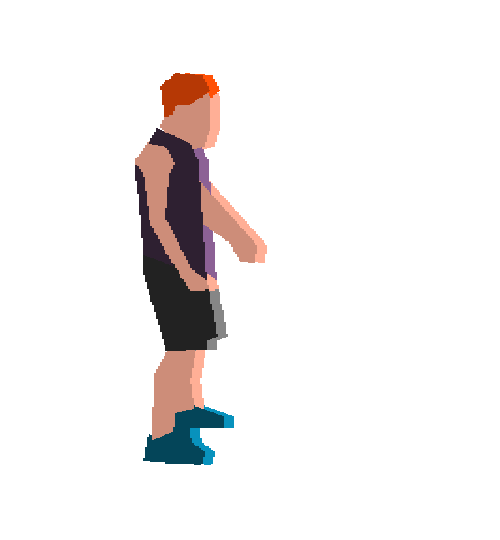} \label{fig:ex1}}
\subfloat{\includegraphics[width=0.60in]{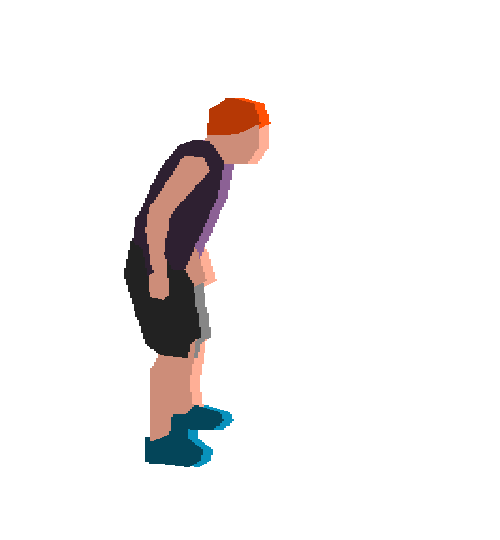} \label{fig:ex1}}
\subfloat{\includegraphics[width=0.60in]{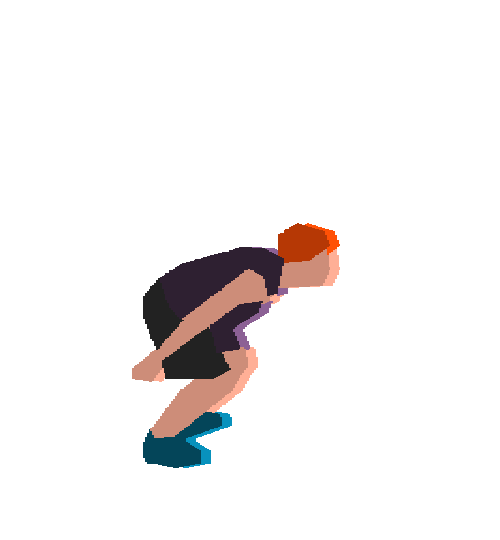} \label{fig:ex1}}
\subfloat{\includegraphics[width=0.60in]{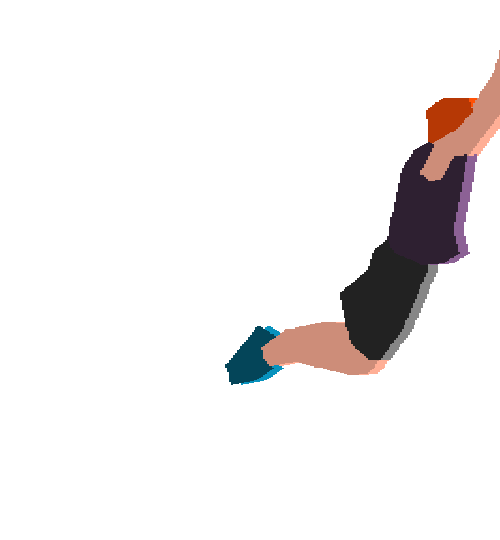} \label{fig:ex1}}\\

\subfloat{\includegraphics[width=0.60in]{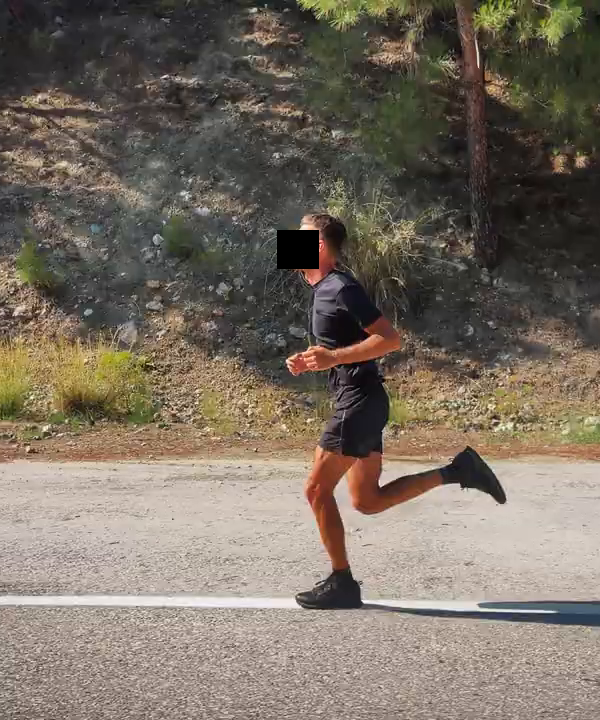} \label{fig:ex1}}
\subfloat{\includegraphics[width=0.60in]{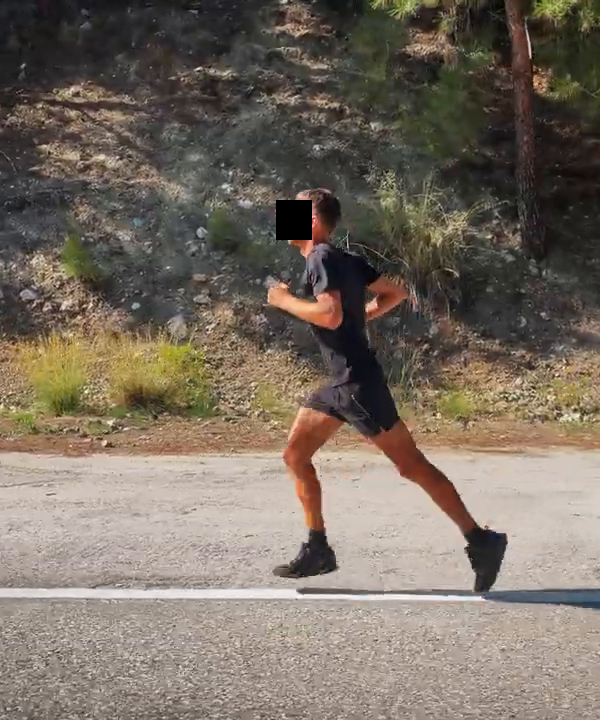} \label{fig:ex1}}
\subfloat{\includegraphics[width=0.60in]{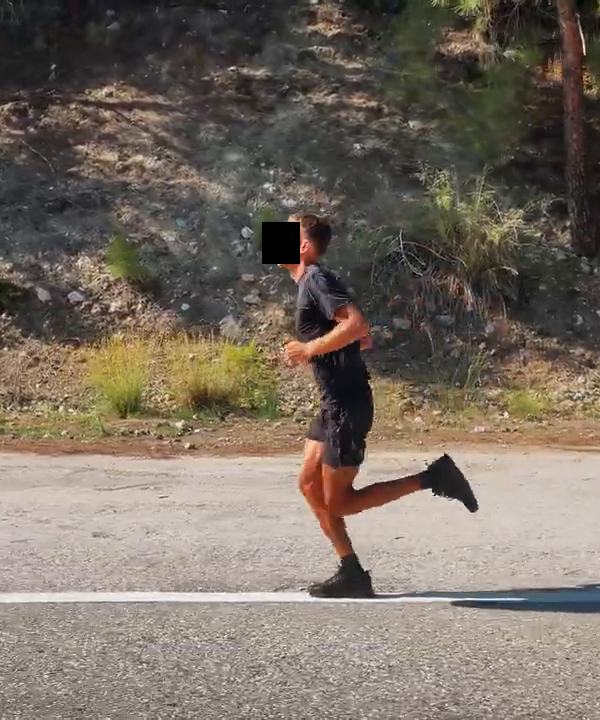} \label{fig:ex1}}
\subfloat{\includegraphics[width=0.60in]{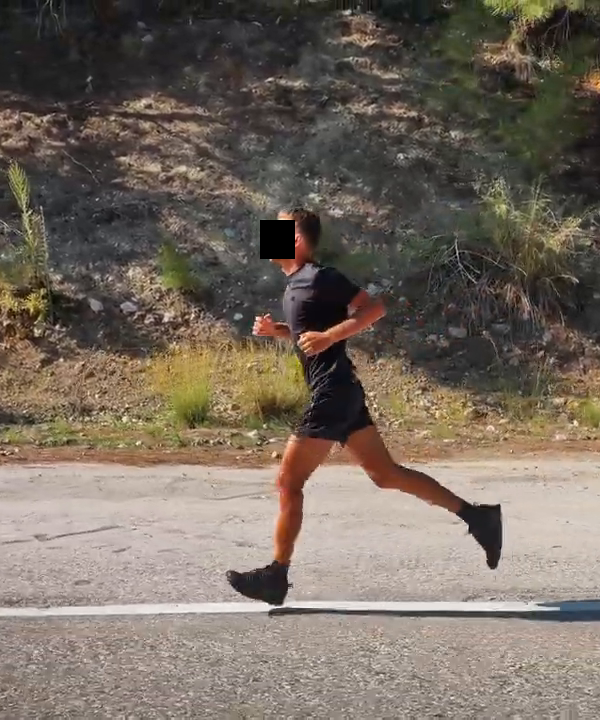} \label{fig:ex1}}
\subfloat{\includegraphics[width=0.60in]{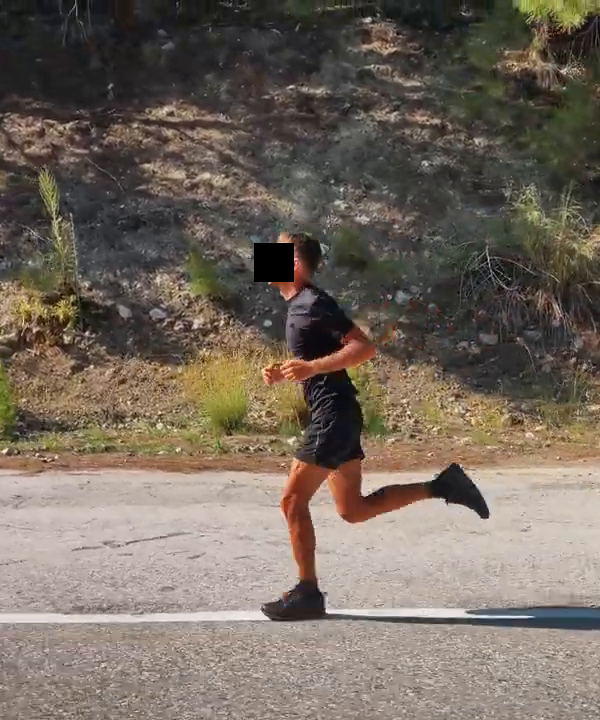} \label{fig:ex1}}\\

\subfloat{\includegraphics[width=0.60in]{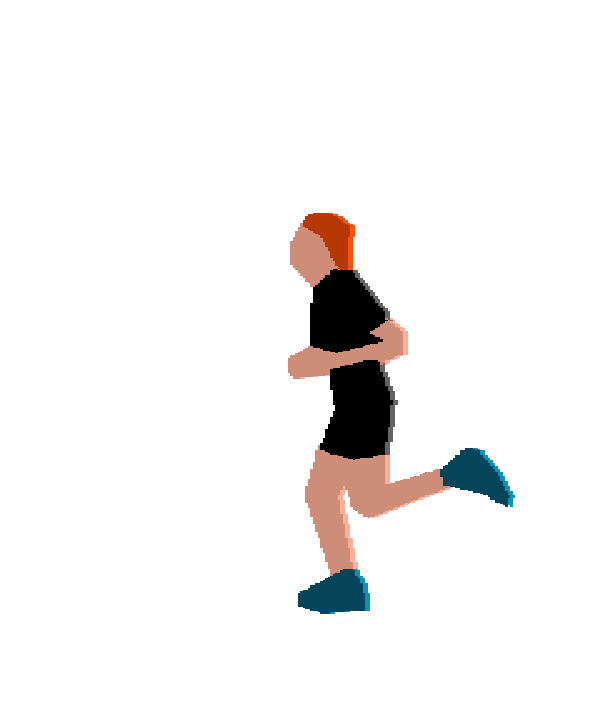} \label{fig:ex1}}
\subfloat{\includegraphics[width=0.60in]{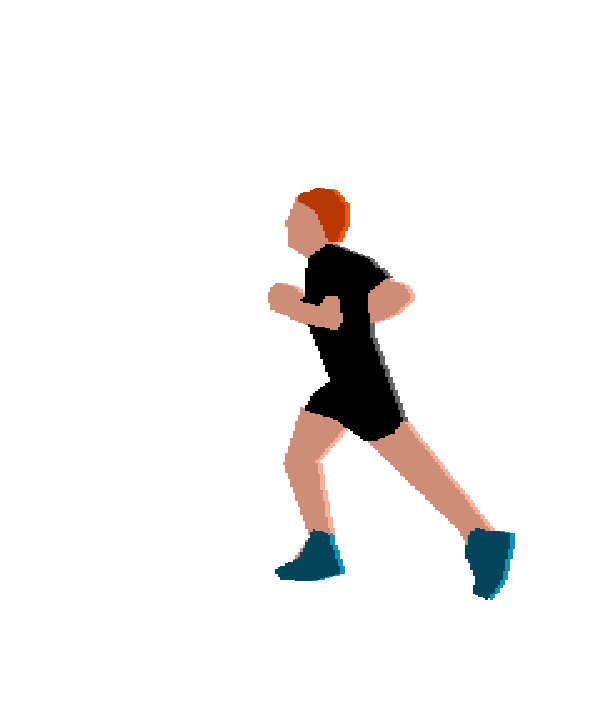} \label{fig:ex1}}
\subfloat{\includegraphics[width=0.60in]{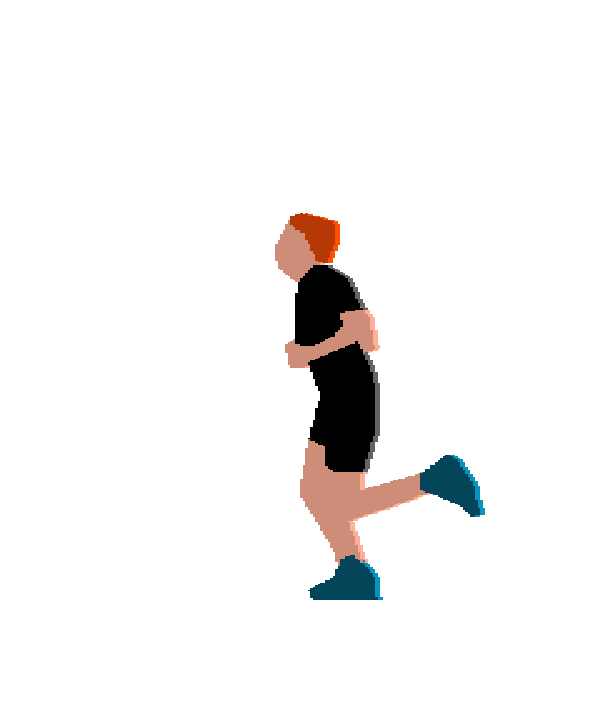} \label{fig:ex1}}
\subfloat{\includegraphics[width=0.60in]{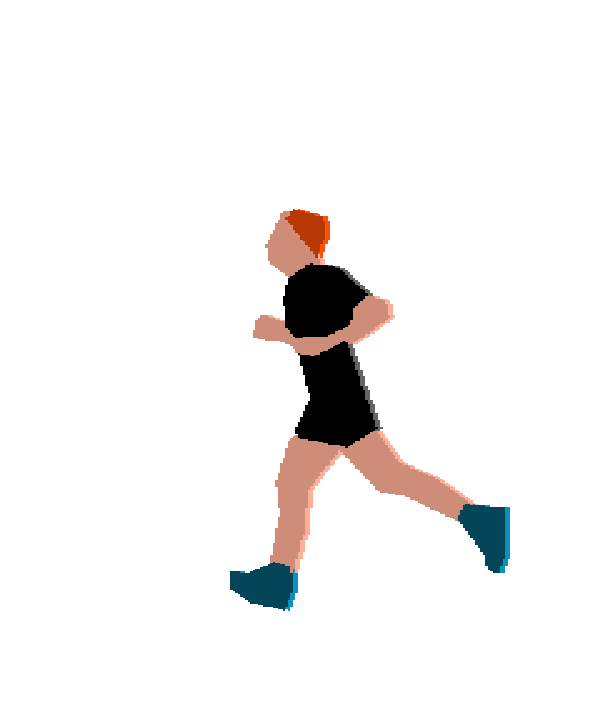} \label{fig:ex1}} 
\subfloat{\includegraphics[width=0.60in]{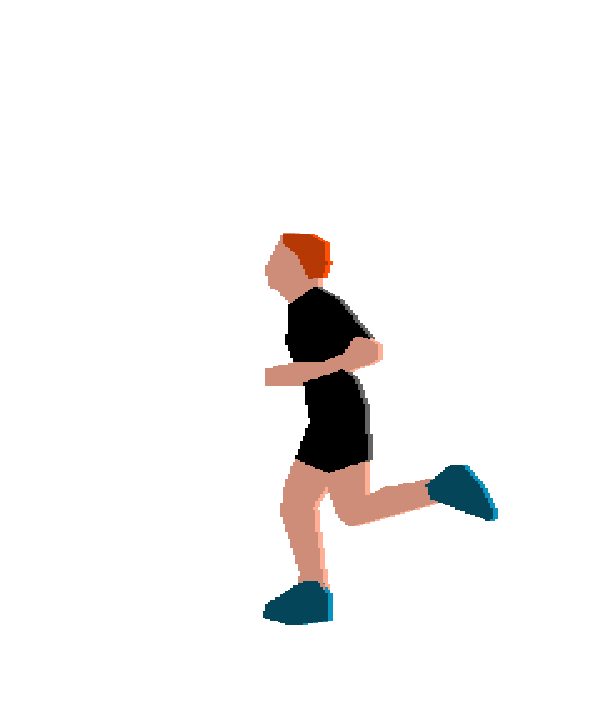} \label{fig:ex1}}\\

\subfloat{\includegraphics[width=0.60in]{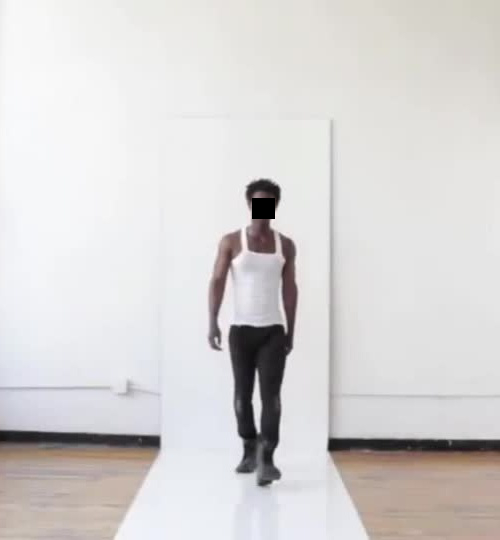} \label{fig:ex1}}
\subfloat{\includegraphics[width=0.60in]{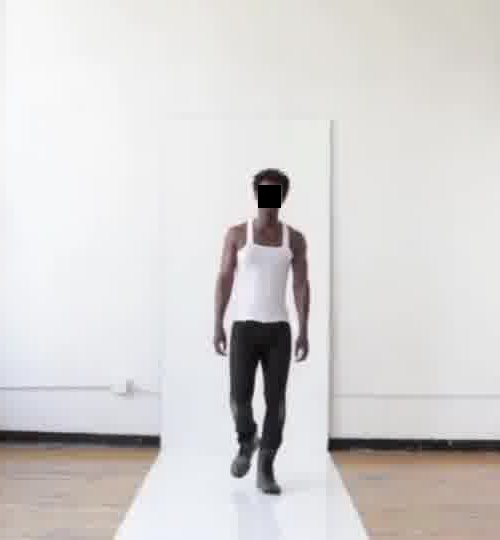} \label{fig:ex1}}
\subfloat{\includegraphics[width=0.60in]{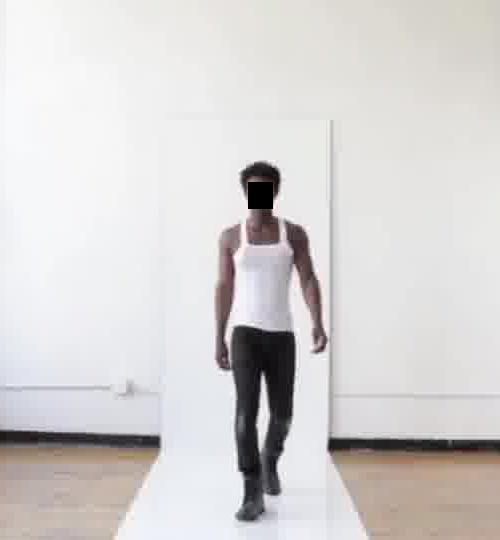} \label{fig:ex1}}
\subfloat{\includegraphics[width=0.60in]{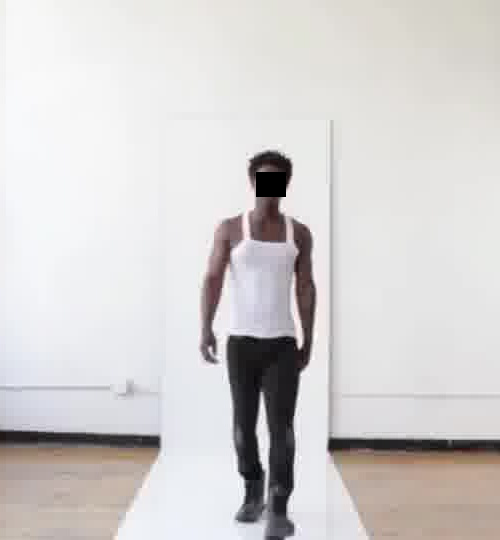} \label{fig:ex1}}
\subfloat{\includegraphics[width=0.60in]{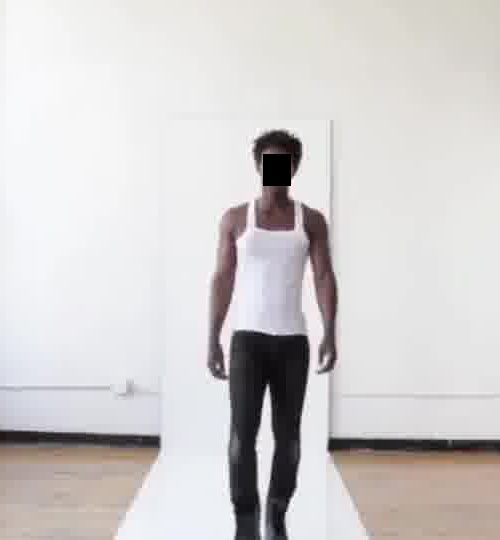} \label{fig:ex1}}\\

\setcounter{subfigure}{0}
\subfloat{\includegraphics[width=0.60in]{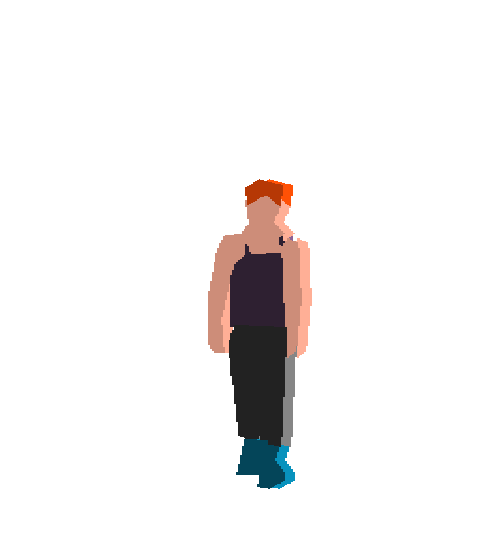} \label{fig:ex1}}
\subfloat{\includegraphics[width=0.60in]{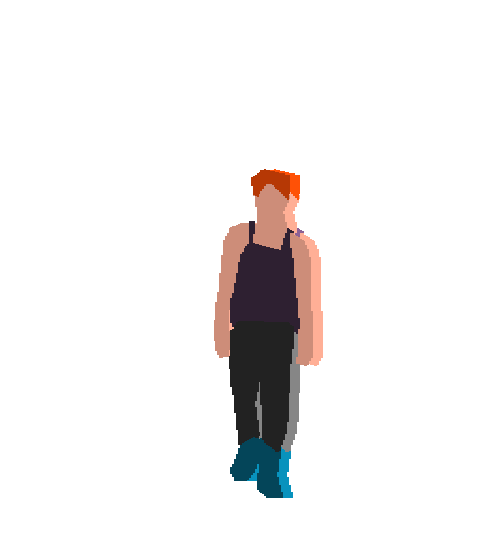} \label{fig:ex1}}
\subfloat{\includegraphics[width=0.60in]{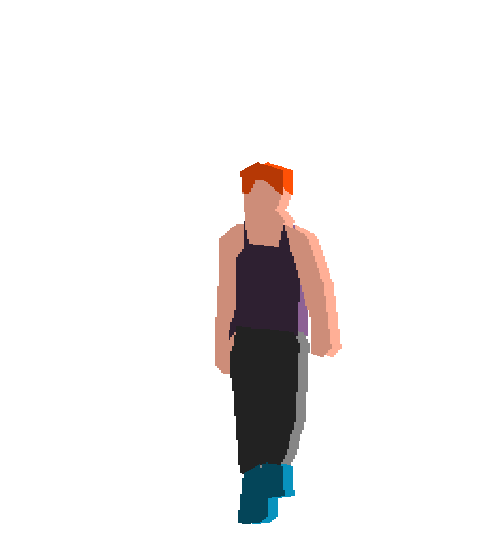} \label{fig:ex1}}
\subfloat{\includegraphics[width=0.60in]{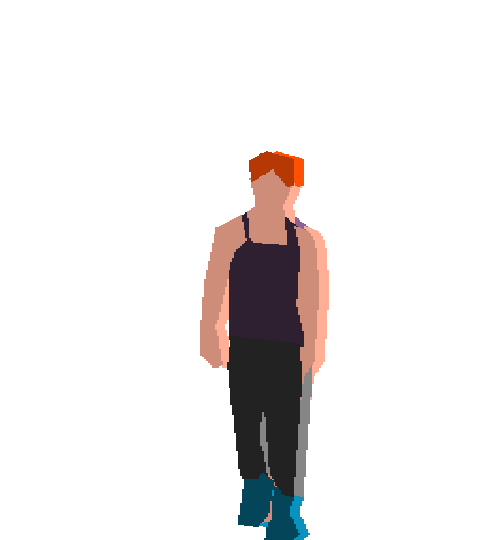} \label{fig:ex1}} 
\subfloat{\includegraphics[width=0.60in]{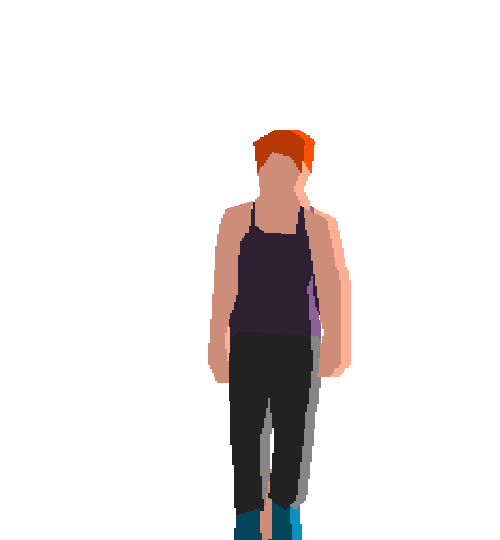} \label{fig:ex1}}\\  

\caption{Example results. Odd rows: input videos (faces have been anonymized); Even rows: method results.}
\label{fig:qualitative1}
\end{figure}

\begin{figure}
\centering
\subfloat{\includegraphics[width=0.60in]{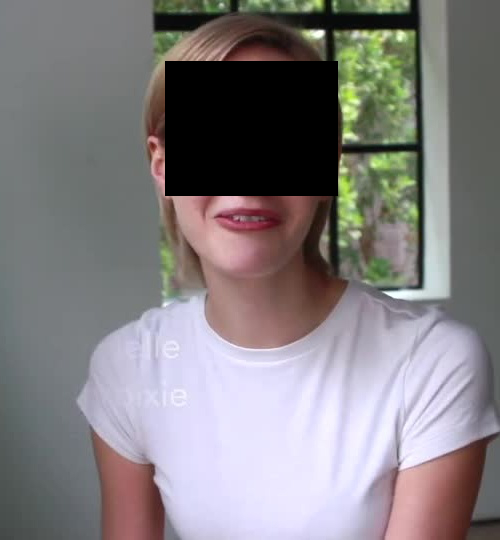} \label{fig:ex1}}
\subfloat{\includegraphics[width=0.60in]{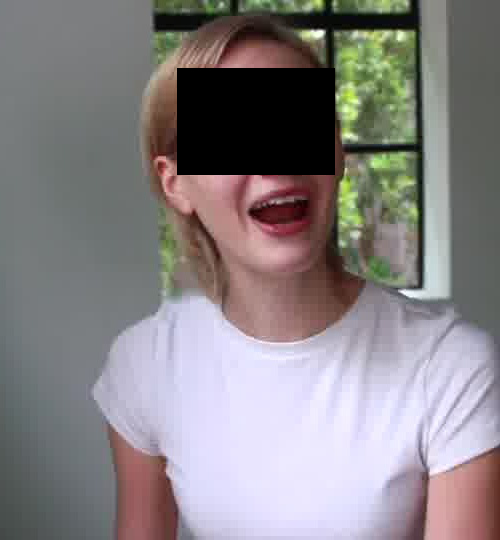} \label{fig:ex1}}
\subfloat{\includegraphics[width=0.60in]{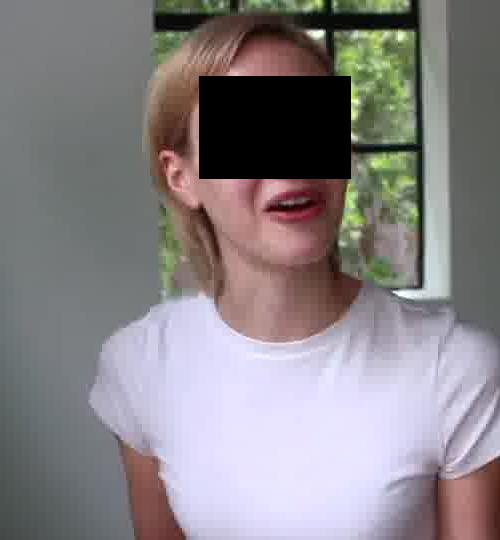} \label{fig:ex1}}
\subfloat{\includegraphics[width=0.60in]{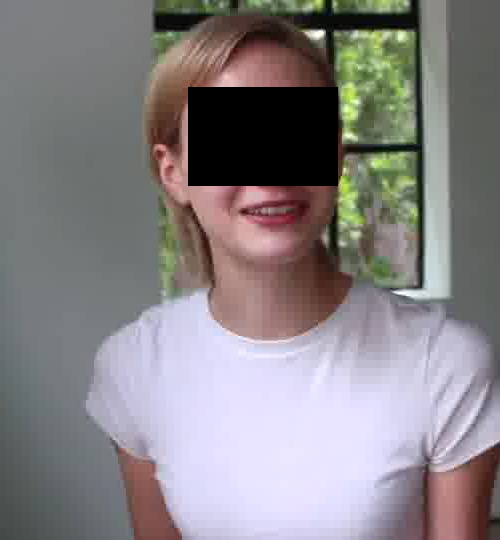} \label{fig:ex1}}
\subfloat{\includegraphics[width=0.60in]{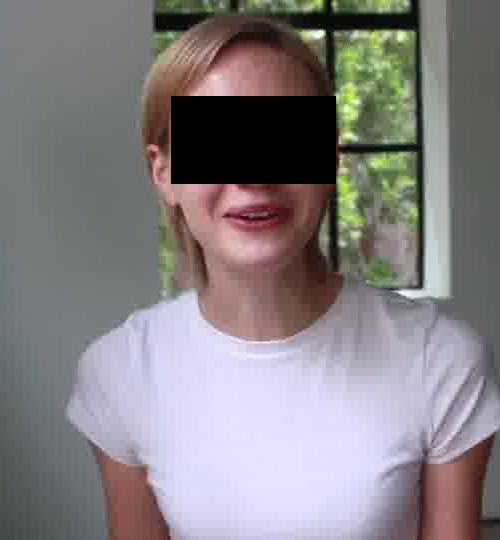} \label{fig:ex1}}\\

\subfloat{\includegraphics[width=0.60in]{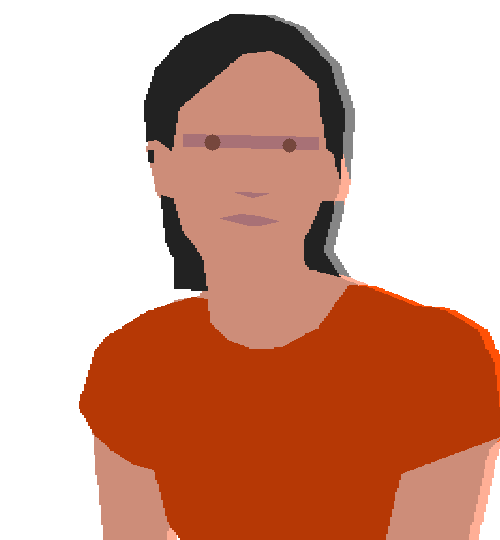} \label{fig:ex1}}
\subfloat{\includegraphics[width=0.60in]{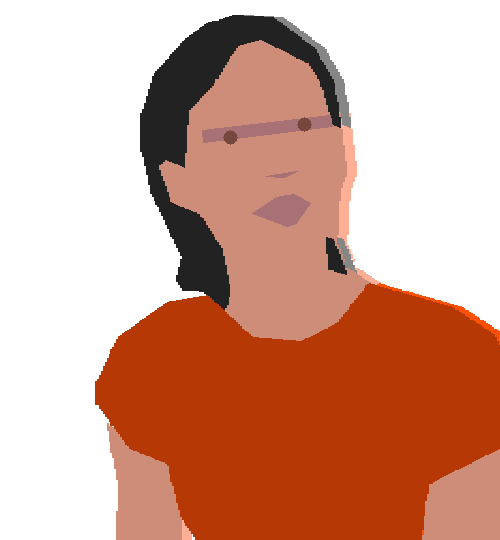} \label{fig:ex1}}
\subfloat{\includegraphics[width=0.60in]{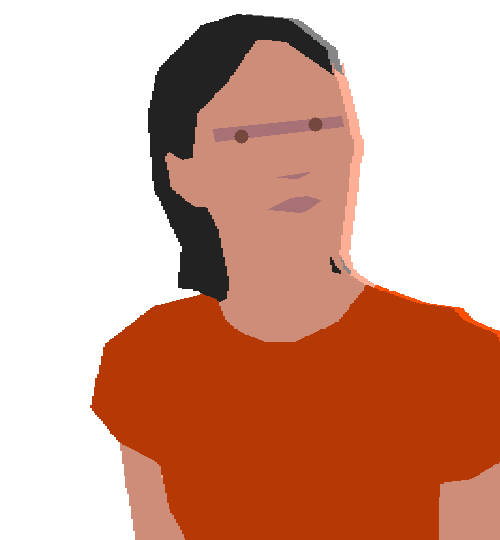} \label{fig:ex1}}
\subfloat{\includegraphics[width=0.60in]{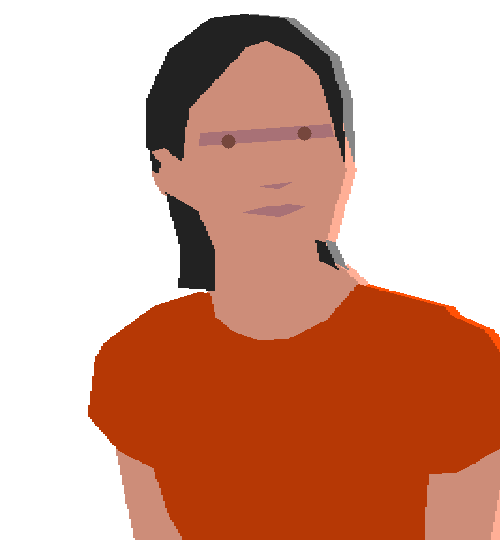} \label{fig:ex1}}
\subfloat{\includegraphics[width=0.60in]{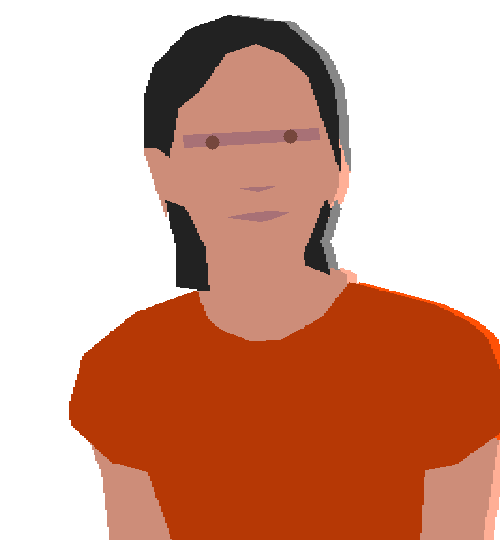} \label{fig:ex1}}\\

\subfloat{\includegraphics[width=0.60in]{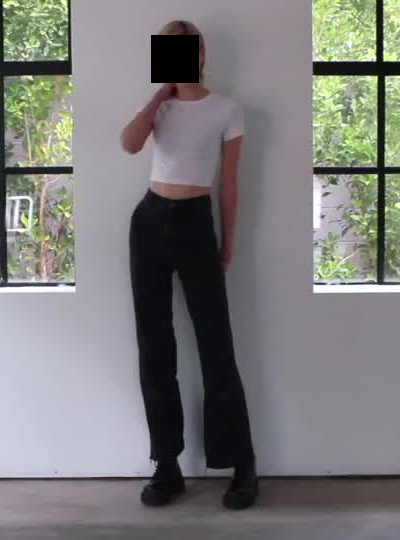} \label{fig:ex1}}   
\subfloat{\includegraphics[width=0.60in]{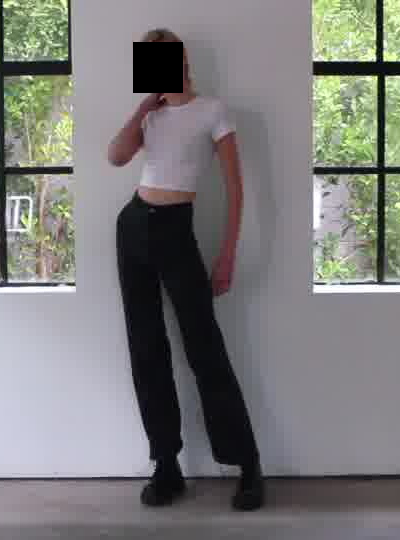} \label{fig:ex1}}
\subfloat{\includegraphics[width=0.60in]{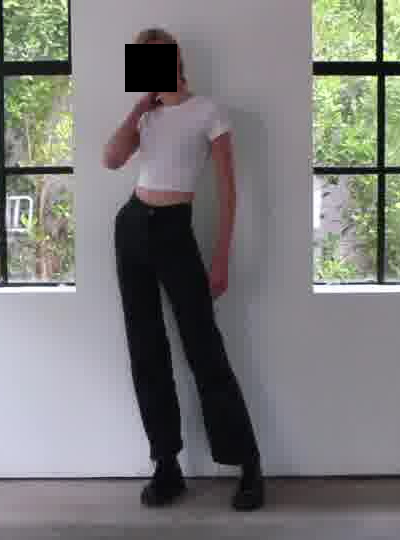} \label{fig:ex1}}
\subfloat{\includegraphics[width=0.60in]{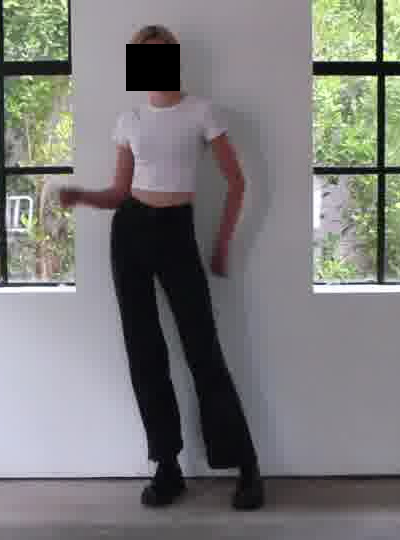} \label{fig:ex1}}
\subfloat{\includegraphics[width=0.60in]{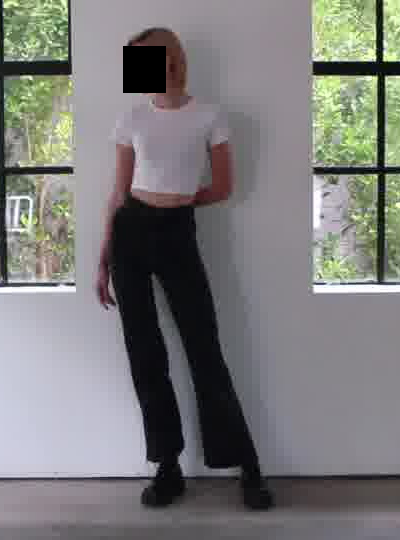} \label{fig:ex1}}\\

\subfloat{\includegraphics[width=0.60in]{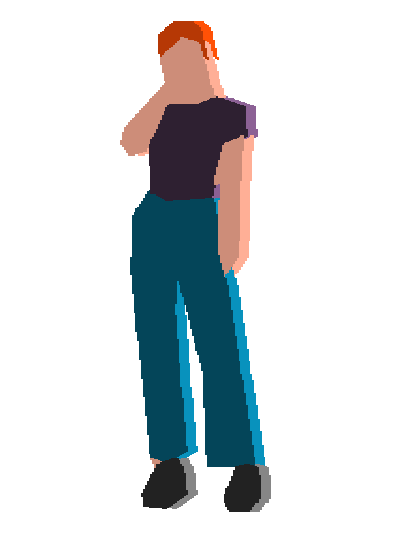} \label{fig:ex1}}
\subfloat{\includegraphics[width=0.60in]{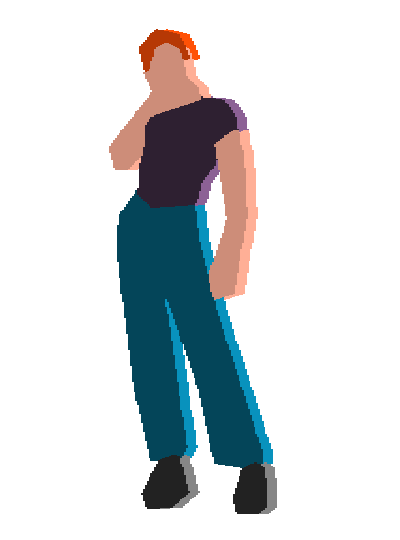} \label{fig:ex1}}
\subfloat{\includegraphics[width=0.60in]{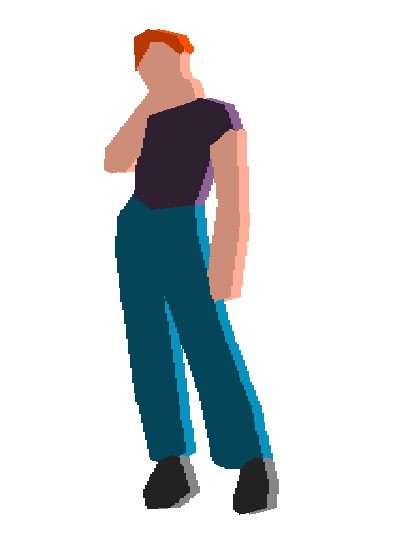} \label{fig:ex1}}
\subfloat{\includegraphics[width=0.60in]{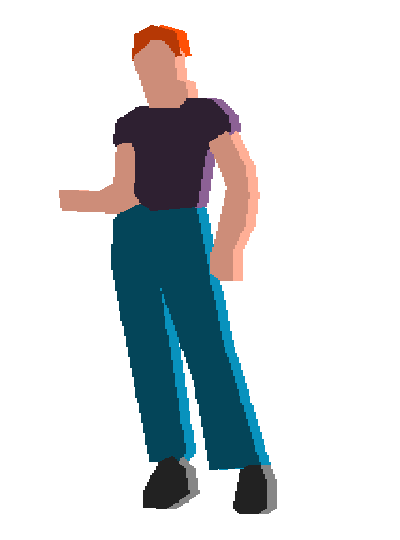} \label{fig:ex1}}
\subfloat{\includegraphics[width=0.60in]{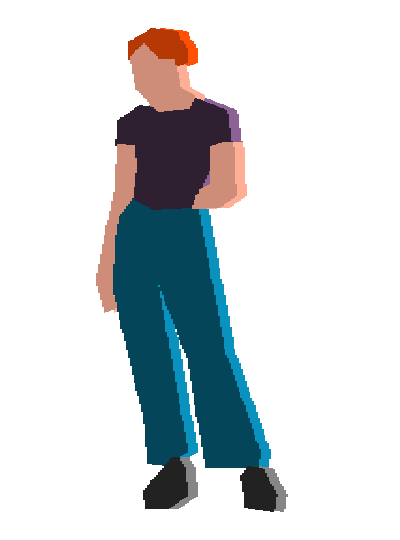} \label{fig:ex1}}\\

\subfloat{\includegraphics[width=0.60in]{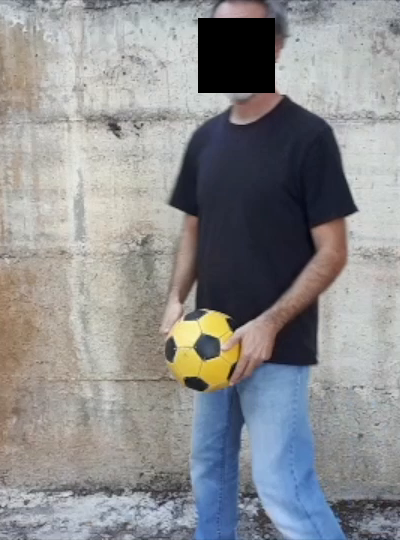} \label{fig:ex1}}
\subfloat{\includegraphics[width=0.60in]{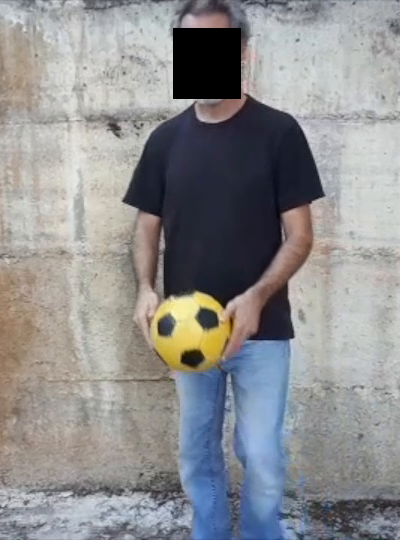} \label{fig:ex1}}
\subfloat{\includegraphics[width=0.60in]{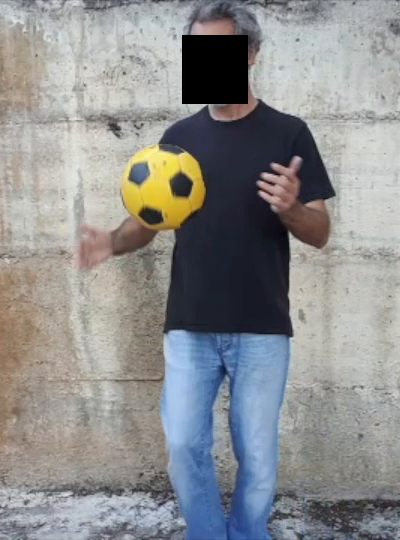} \label{fig:ex1}}
\subfloat{\includegraphics[width=0.60in]{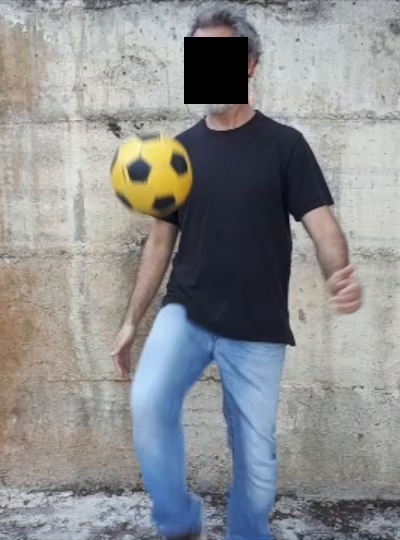} \label{fig:ex1}}
\subfloat{\includegraphics[width=0.60in]{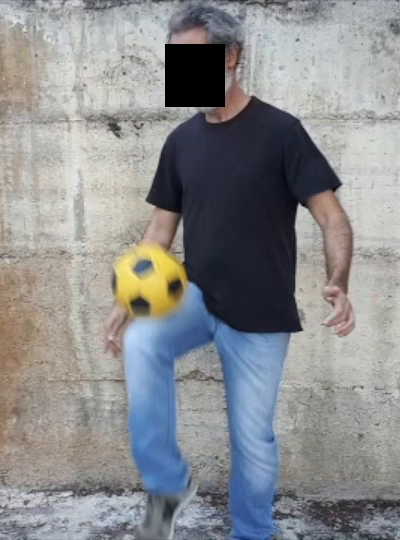} \label{fig:ex1}}\\

\setcounter{subfigure}{0} 
\subfloat{\includegraphics[width=0.60in]{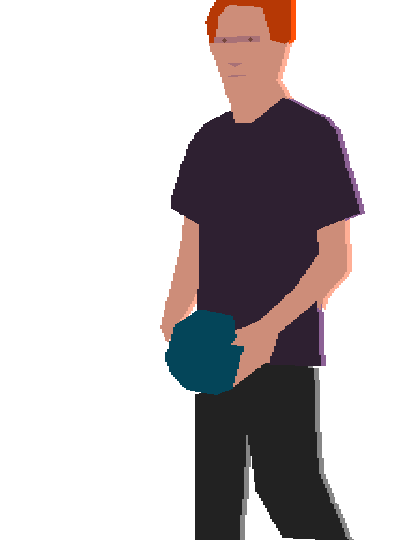} \label{fig:ex1}}
\subfloat{\includegraphics[width=0.60in]{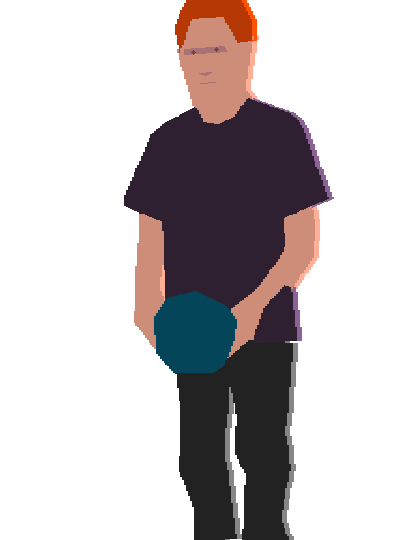} \label{fig:ex1}}
\subfloat{\includegraphics[width=0.60in]{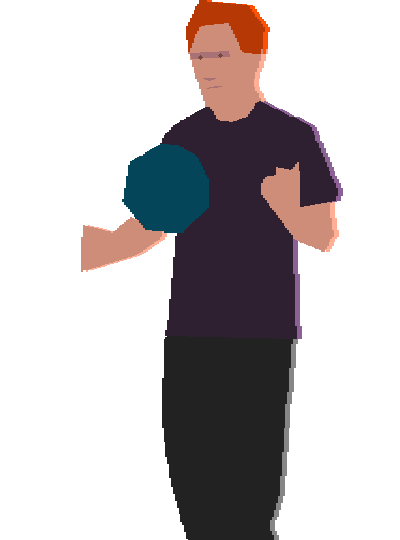} \label{fig:ex1}}
\subfloat{\includegraphics[width=0.60in]{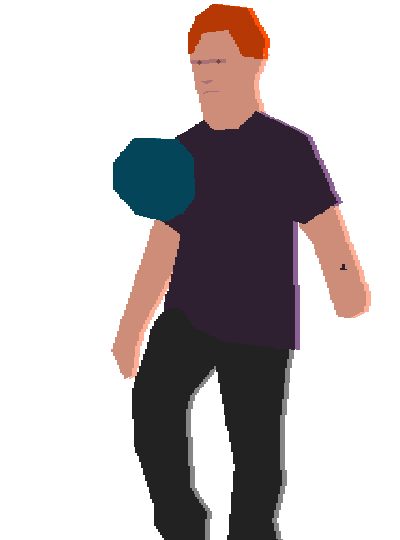} \label{fig:ex1}} 
\subfloat{\includegraphics[width=0.60in]{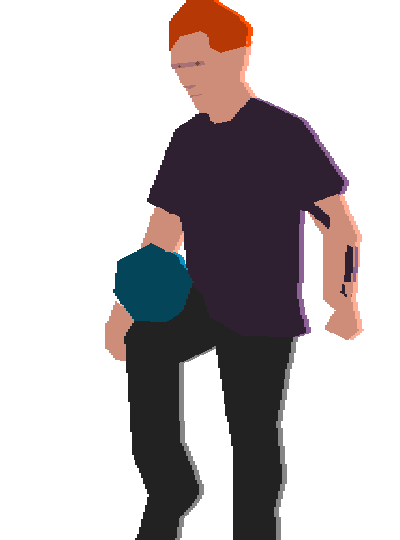} \label{fig:ex1}}\\  

\caption{Example results. Odd rows: input videos (faces have been anonymized); Even rows: method results.} 
\label{fig:qualitative2}    
\end{figure}

In order to evaluate the performance of the method, we collected a set of 25 short videos ranging from 5 to 20 seconds in duration. The videos showcase a lone human subject displaying various clothing styles and engaging in diverse actions. The videos were obtained from the UCF101 Human Actions dataset \cite{ucf101-dataset}, the Fashion Video Dataset \cite{fashion-video-dataset}, the Kinetics dataset \cite{kinetics-dataset} and YouTube. The assembled datased includes both low and hight resolution videos and both in-the-wild and controlled-setting videos. The dataset and the results of the experiments can be found in \cite{lester}. Figures ~\ref{fig:qualitative1} and ~\ref{fig:qualitative2} shows some of example results. 

The subjective quality assessment of the results is summarized in Table ~\ref{table:results1}. We report the MOS (mean opinion score described in ITU-T P.800.1 \cite{p800}) of the ACR (absolute category rating described in ITU-T P.910 recommendation \cite{p910} scale from 1=Bad to 5=Excellent) of 3 aspects: shape quality, color consistency and temporal consistency, plus and overall score. The videos are grouped into three categories, in-the-wild low resolution (ITW-LR), in-the-wild high resolution (ITW-HR) and controlled setting high resolution (C-HR).\\ 

\begin{table} 
\centering
\begin{tabular}{lccccc} \toprule
 subset & \# & shape & color & temporal & overall \\ \midrule
 ITW-LR\textsuperscript{a} & 5 & 3.20 & 3.24 & 3.84 & 3.36 \\
 ITW-HR\textsuperscript{b} & 14 & 3.81 & 4.03 & 4.33 & 3.99 \\
 C-HR\textsuperscript{c} & 6 & 4.70 & 4.60 & 4.67 & 4.67 \\\midrule
 \textbf{all} & & \textbf{3.90} & \textbf{4.01} & \textbf{4.31} & \textbf{4.02} \\ \bottomrule
\end{tabular}
\caption{Subjective quality assessment MOS values of the ACR (1=Bad to 5=Excellent) of shape quality, color consistency, temporal consistency and overall quality. \textsuperscript{a}ITW-LR $\rightarrow$ in-the-wild low resolution; \textsuperscript{b}ITW-HR $\rightarrow$ in-the-wild high resolution; \textsuperscript{c}C-HR $\rightarrow$ controlled setting high resolution}
\label{table:results1}
\end{table}

\noindent
\textbf{Strengths:}
The results show that the method is able to synthetize quality animations from diverse input videos (an average MOE 4.02 over 5). The performance significantly improves (4.67) on high resolution videos taken in controlled conditions (videos 2, 11, 12, 13, 14 and 15). The method exhibits a very good temporal consistency and can correctly process videos with different poses and appearances, dynamic shots (e.g. video 5), videos with partial shots (e.g. the medium close-up shot in video 7) and diverse backgrounds.\\ 

\begin{figure}
\centering
\subfloat{\includegraphics[width=0.80in]{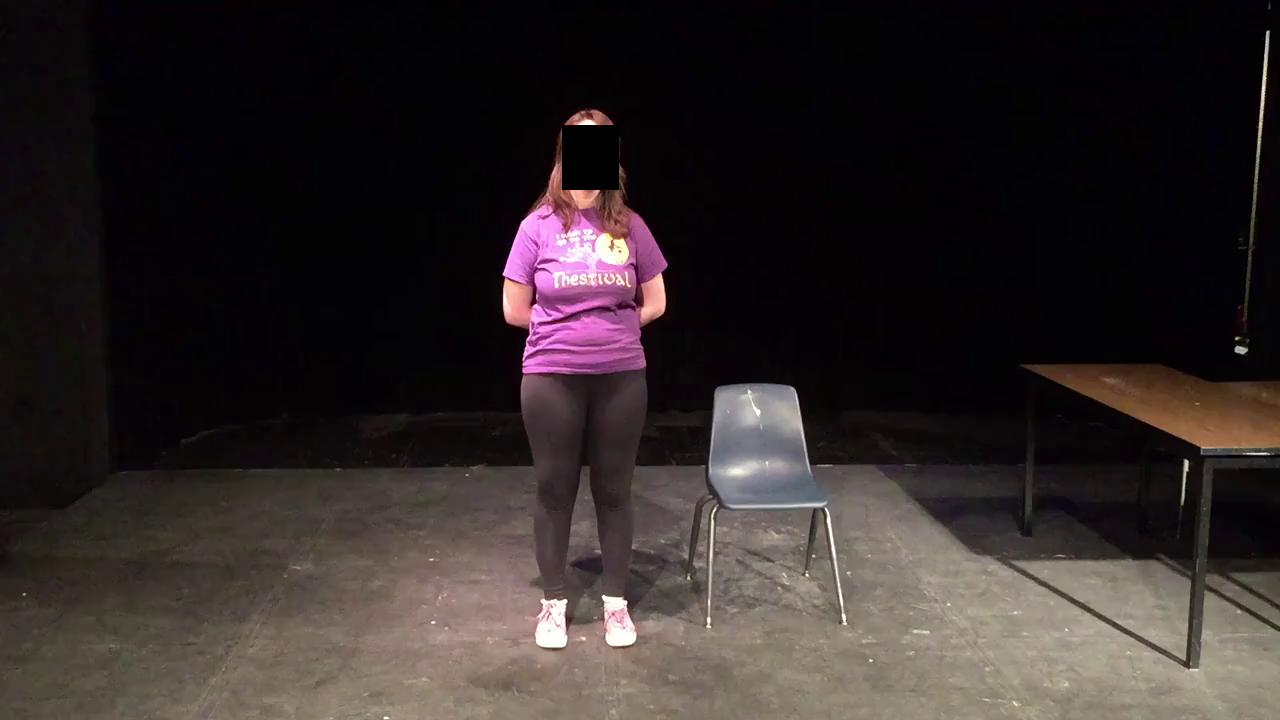} \label{fig:ex1}}
\subfloat{\includegraphics[width=0.80in]{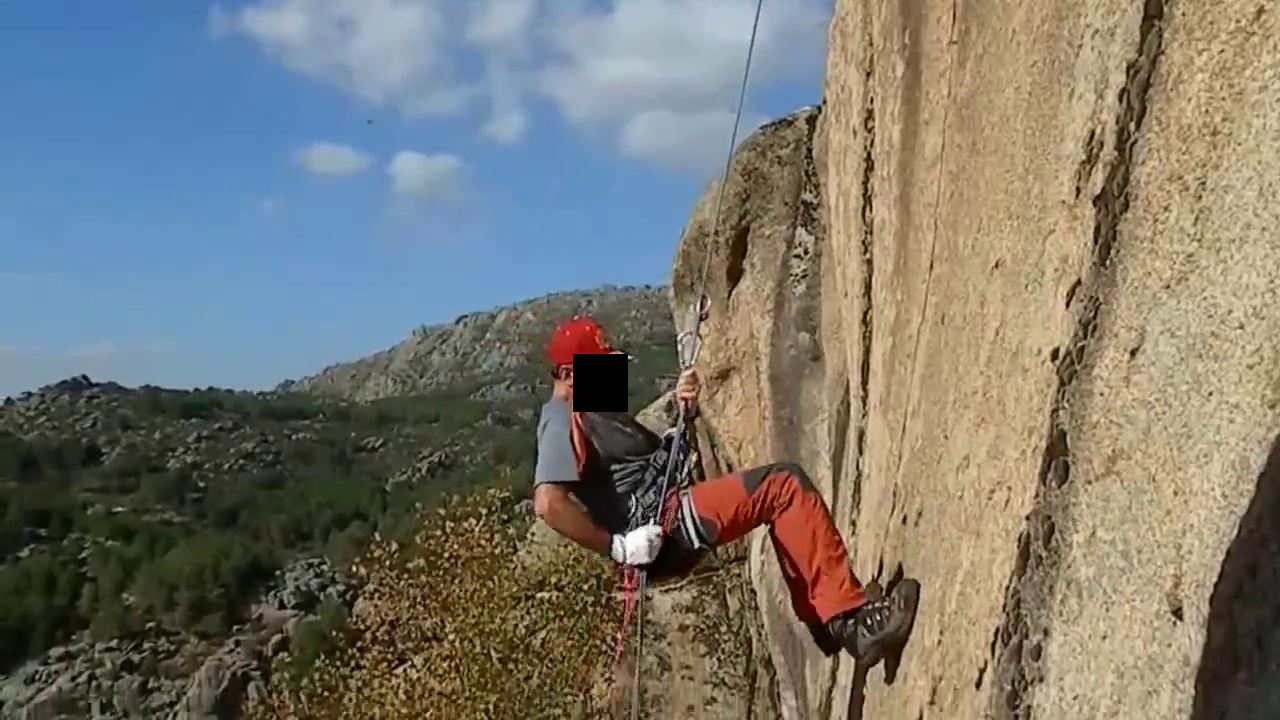} \label{fig:ex1}}
\subfloat{\includegraphics[width=0.80in]{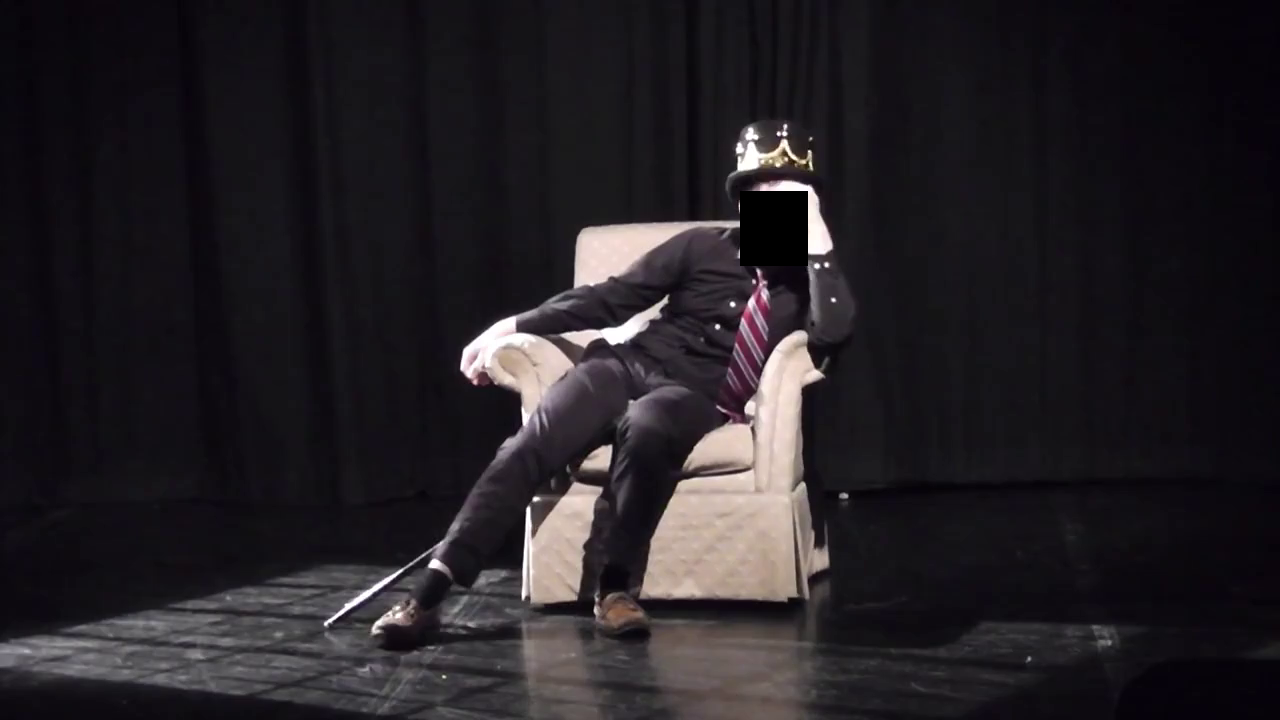} \label{fig:ex1}}
\subfloat{\includegraphics[width=0.80in]{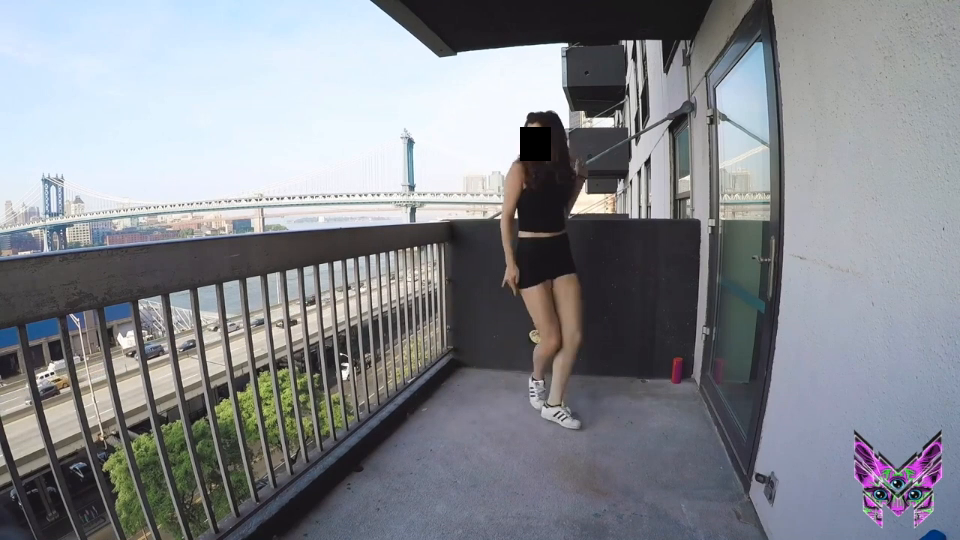} \label{fig:ex1}}\\
\subfloat{\includegraphics[width=0.80in]{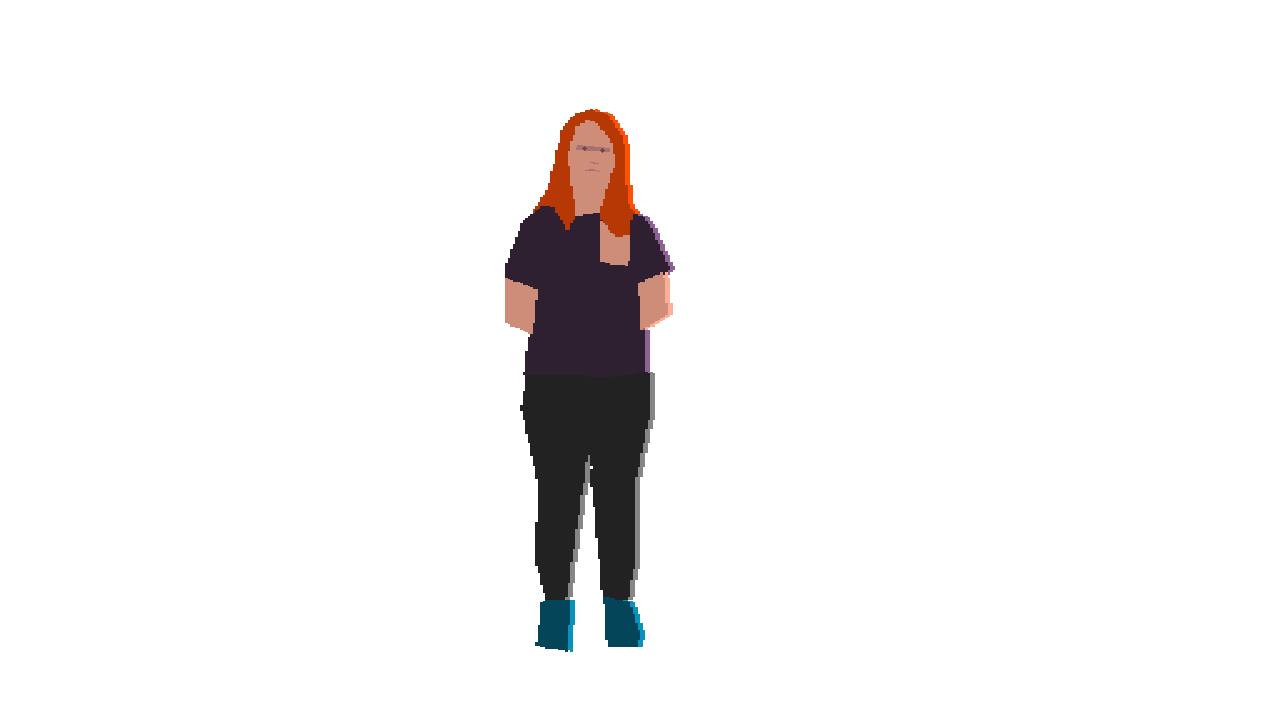} \label{fig:ex1}}
\subfloat{\includegraphics[width=0.80in]{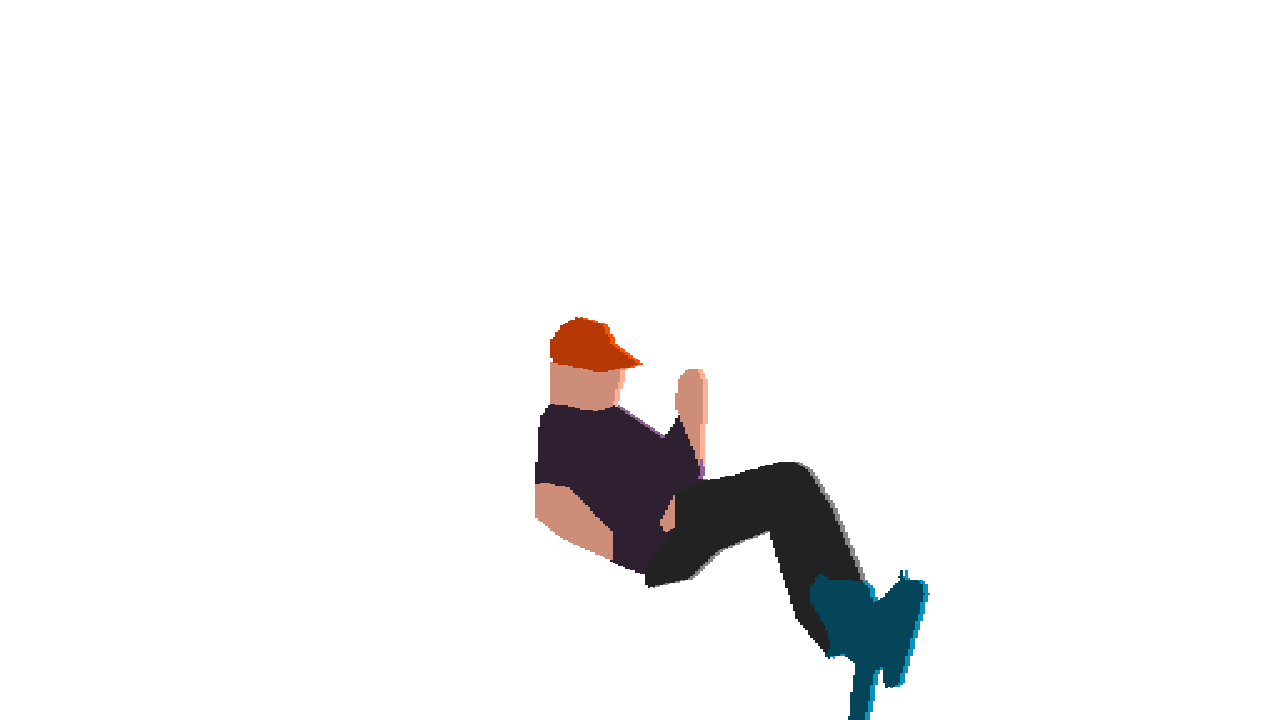} \label{fig:ex1}}
\subfloat{\includegraphics[width=0.80in]{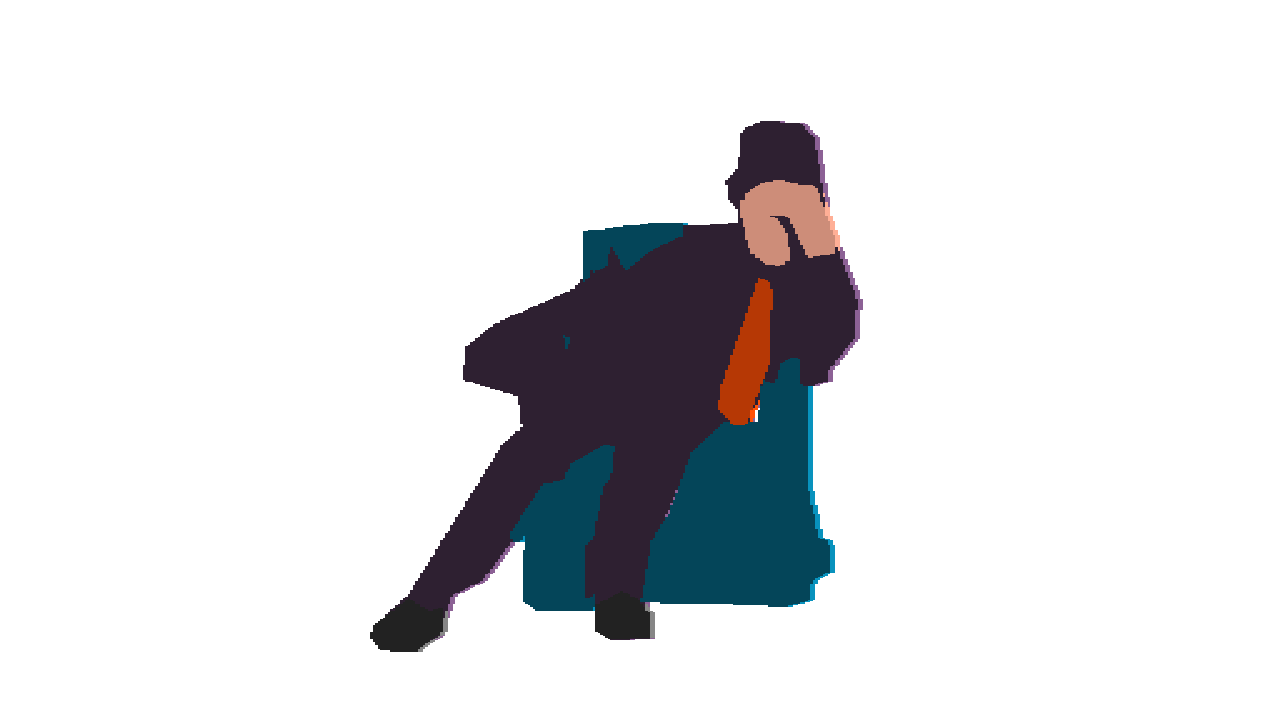} \label{fig:ex1}} 
\subfloat{\includegraphics[width=0.80in]{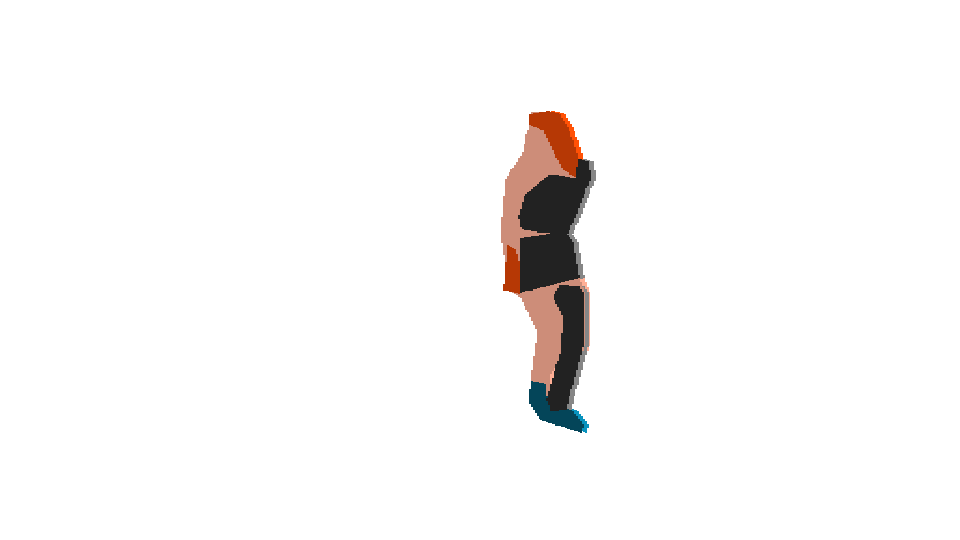} \label{fig:ex1}} 

\caption{Example failure cases. Top row: input videos (videos 22, 23, 25 and 8 from the test dataset, faces have been anonymized); bottom row: results.} 
\label{fig:failures}  
\end{figure}

\noindent
\textbf{Limitations:}
The performance decreases in videos with low resolution (all videos from the UCF101 Human Actions dataset), complex backgrounds (videos 17 and 20), low contrast colors (videos 8, 10, 22 and 23), inappropriate lighting (videos 14 and 25) and occlusions (video 23). Figure ~\ref{fig:failures} shows four example frames with errors from four different videos (videos 22, 23, 25 and 8 from the test dataset).\\

\noindent
\textbf{Computational cost:}
The experiments have been run on a high-end server with a quadcore Intel i7-3820 at 3.6 GHz with 64 GB of DDR3 RAM, and 4 NVIDIA Tesla K40 GPU cards with 12 GB of GDDR5 each. For a single frame, step 1 (segmentation and tracking) takes 0.56 seconds, step 2 (contours simplification) takes 0.43 seconds and step 3 (finishing details, mainly facial features) takes 1.32 seconds. Overall the method operates at 0.43 FPS (1 FPS without facial features).\\ 

\noindent
\textbf{Discussion:}
The results obtained are very satisfactory and validate the base hypothesis, that the latest advances in segmentation and tracking make this technique an effective method for generating rotoscopic animations from videos. The limitations detected do not invalidate the initial hypothesis since it is expected that the method will be used on selected or purposely created videos. Nor does its computational cost since the method is expected to be used in editing tasks that do not require real-time execution. The method is primarily designed for videos featuring a single person, but could be adapted to videos with multiple people by using full-body masks and multiple parameterizations for segmentation prompts and color palettes.\\

\section{Conclusions}
In this work we present Lester, a method for automatically synthetise 2D retro-style animated characters from videos. Rather than addressing the challenge with an image-to-image generative model, or with human pose estimation tools, it is addressed mainly as an object segmentation and tracking problem. Subjective quality assessment shows that the approach is feasible and that the method can generate quality animations from videos with different poses and appearances, dynamic shots, partial shots and diverse backgrounds. Although it is not the objective, it is worth mentioning that the performance of the method is less consistent in-the-wild, especially against low resolution videos. But the method is expected to be used on selected or custom-made videos in semi-controlled environments.

Due to its capabilities, the method constitutes an effective tool that significantly reduces the cost of generating rotoscopic animations for video games or animated films. Due to its characteristics, it is more simple and deterministic than generative models based pipelines and more practical and flexible than techniques based on human pose estimation.

section*{Acknowledgements}
This work is partially supported by the Spanish Ministry of Science and Innovation under contract PID2019-107255GB, and by the SGR programme 2021-SGR-00478 of the Catalan Government.  


\section*{Declarations}

\subsection*{Competing interests}
The authors declare that they have no competing interests.

\subsection*{Data availability }
The dataset and the results of the experiments are publicly available at \url{https://github.com/rtous/lester}.  




\begin{thebibliography}{35}
\ifx \bisbn   \undefined \def \bisbn  #1{ISBN #1}\fi
\ifx \binits  \undefined \def \binits#1{#1}\fi
\ifx \bauthor  \undefined \def \bauthor#1{#1}\fi
\ifx \batitle  \undefined \def \batitle#1{#1}\fi
\ifx \bjtitle  \undefined \def \bjtitle#1{#1}\fi
\ifx \bvolume  \undefined \def \bvolume#1{\textbf{#1}}\fi
\ifx \byear  \undefined \def \byear#1{#1}\fi
\ifx \bissue  \undefined \def \bissue#1{#1}\fi
\ifx \bfpage  \undefined \def \bfpage#1{#1}\fi
\ifx \blpage  \undefined \def \blpage #1{#1}\fi
\ifx \burl  \undefined \def \burl#1{\textsf{#1}}\fi
\ifx \doiurl  \undefined \def \doiurl#1{\url{https://doi.org/#1}}\fi
\ifx \betal  \undefined \def \betal{\textit{et al.}}\fi
\ifx \binstitute  \undefined \def \binstitute#1{#1}\fi
\ifx \binstitutionaled  \undefined \def \binstitutionaled#1{#1}\fi
\ifx \bctitle  \undefined \def \bctitle#1{#1}\fi
\ifx \beditor  \undefined \def \beditor#1{#1}\fi
\ifx \bpublisher  \undefined \def \bpublisher#1{#1}\fi
\ifx \bbtitle  \undefined \def \bbtitle#1{#1}\fi
\ifx \bedition  \undefined \def \bedition#1{#1}\fi
\ifx \bseriesno  \undefined \def \bseriesno#1{#1}\fi
\ifx \blocation  \undefined \def \blocation#1{#1}\fi
\ifx \bsertitle  \undefined \def \bsertitle#1{#1}\fi
\ifx \bsnm \undefined \def \bsnm#1{#1}\fi
\ifx \bsuffix \undefined \def \bsuffix#1{#1}\fi
\ifx \bparticle \undefined \def \bparticle#1{#1}\fi
\ifx \barticle \undefined \def \barticle#1{#1}\fi
\bibcommenthead
\ifx \bconfdate \undefined \def \bconfdate #1{#1}\fi
\ifx \botherref \undefined \def \botherref #1{#1}\fi
\ifx \url \undefined \def \url#1{\textsf{#1}}\fi
\ifx \bchapter \undefined \def \bchapter#1{#1}\fi
\ifx \bbook \undefined \def \bbook#1{#1}\fi
\ifx \bcomment \undefined \def \bcomment#1{#1}\fi
\ifx \oauthor \undefined \def \oauthor#1{#1}\fi
\ifx \citeauthoryear \undefined \def \citeauthoryear#1{#1}\fi
\ifx \endbibitem  \undefined \def \endbibitem {}\fi
\ifx \bconflocation  \undefined \def \bconflocation#1{#1}\fi
\ifx \arxivurl  \undefined \def \arxivurl#1{\textsf{#1}}\fi
\csname PreBibitemsHook\endcsname

\bibitem[\protect\citeauthoryear{Kirillov et~al.}{2023}]{sam}
\begin{botherref}
\oauthor{\bsnm{Kirillov}, \binits{A.}},
\oauthor{\bsnm{Mintun}, \binits{E.}},
\oauthor{\bsnm{Ravi}, \binits{N.}},
\oauthor{\bsnm{Mao}, \binits{H.}},
\oauthor{\bsnm{Rolland}, \binits{C.}},
\oauthor{\bsnm{Gustafson}, \binits{L.}},
\oauthor{\bsnm{Xiao}, \binits{T.}},
\oauthor{\bsnm{Whitehead}, \binits{S.}},
\oauthor{\bsnm{Berg}, \binits{A.C.}},
\oauthor{\bsnm{Lo}, \binits{W.-Y.}}, et al.:
Segment anything.
arXiv preprint arXiv:2304.02643
(2023)
\end{botherref}
\endbibitem

\bibitem[\protect\citeauthoryear{Cheng et~al.}{2023}]{sam-track}
\begin{botherref}
\oauthor{\bsnm{Cheng}, \binits{Y.}},
\oauthor{\bsnm{Li}, \binits{L.}},
\oauthor{\bsnm{Xu}, \binits{Y.}},
\oauthor{\bsnm{Li}, \binits{X.}},
\oauthor{\bsnm{Yang}, \binits{Z.}},
\oauthor{\bsnm{Wang}, \binits{W.}},
\oauthor{\bsnm{Yang}, \binits{Y.}}:
Segment and track anything.
arXiv preprint arXiv:2305.06558
(2023)
\end{botherref}
\endbibitem

\bibitem[\protect\citeauthoryear{Douglas and Peucker}{2011}]{douglas-peucker}
\begin{bbook}
\bauthor{\bsnm{Douglas}, \binits{D.H.}},
\bauthor{\bsnm{Peucker}, \binits{T.K.}}:
\bbtitle{{Algorithms for the Reduction of the Number of Points Required to
  Represent a Digitized Line or its Caricature}},
pp. \bfpage{15}--\blpage{28}
(\byear{2011}).
\doiurl{10.1002/9780470669488.ch2}
\end{bbook}
\endbibitem

\bibitem[\protect\citeauthoryear{Chen et~al.}{2018}]{cartoongan}
\begin{bchapter}
\bauthor{\bsnm{Chen}, \binits{Y.}},
\bauthor{\bsnm{Lai}, \binits{Y.-K.}},
\bauthor{\bsnm{Liu}, \binits{Y.-J.}}:
\bctitle{Cartoongan: Generative adversarial networks for photo cartoonization.}
In: \bbtitle{CVPR},
pp. \bfpage{9465}--\blpage{9474}.
\bpublisher{IEEE Computer Society}, \blocation{???}
(\byear{2018})
\end{bchapter}
\endbibitem

\bibitem[\protect\citeauthoryear{Chen et~al.}{2020}]{animegan}
\begin{bbook}
\bauthor{\bsnm{Chen}, \binits{J.}},
\bauthor{\bsnm{Liu}, \binits{G.}},
\bauthor{\bsnm{Chen}, \binits{X.}}:
\bbtitle{AnimeGAN: A Novel Lightweight GAN for Photo Animation},
pp. \bfpage{242}--\blpage{256}.
\bpublisher{Artificial Intelligence Algorithms and Applications},
  \blocation{???}
(\byear{2020}).
\doiurl{10.1007/978-981-15-5577-0\_18}
\end{bbook}
\endbibitem

\bibitem[\protect\citeauthoryear{Liu et~al.}{2022}]{cartoonizationLiu2022}
\begin{botherref}
\oauthor{\bsnm{Liu}, \binits{Z.}},
\oauthor{\bsnm{Li}, \binits{L.}},
\oauthor{\bsnm{Jiang}, \binits{H.}},
\oauthor{\bsnm{Jin}, \binits{X.}},
\oauthor{\bsnm{Tu}, \binits{D.}},
\oauthor{\bsnm{Wang}, \binits{S.}},
\oauthor{\bsnm{Zha}, \binits{Z.}}:
Unsupervised coherent video cartoonization with perceptual motion consistency.
CoRR
\textbf{abs/2204.00795}
(2022)
\doiurl{10.48550/arXiv.2204.00795}
{\href{https://arxiv.org/abs/2204.00795}{{2204.00795}}}
\end{botherref}
\endbibitem

\bibitem[\protect\citeauthoryear{Rombach et~al.}{2021}]{stablediffusion}
\begin{botherref}
\oauthor{\bsnm{Rombach}, \binits{R.}},
\oauthor{\bsnm{Blattmann}, \binits{A.}},
\oauthor{\bsnm{Lorenz}, \binits{D.}},
\oauthor{\bsnm{Esser}, \binits{P.}},
\oauthor{\bsnm{Ommer}, \binits{B.}}:
High-Resolution Image Synthesis with Latent Diffusion Models
(2021)
\end{botherref}
\endbibitem

\bibitem[\protect\citeauthoryear{Meng et~al.}{2022}]{image2image}
\begin{bchapter}
\bauthor{\bsnm{Meng}, \binits{C.}},
\bauthor{\bsnm{He}, \binits{Y.}},
\bauthor{\bsnm{Song}, \binits{Y.}},
\bauthor{\bsnm{Song}, \binits{J.}},
\bauthor{\bsnm{Wu}, \binits{J.}},
\bauthor{\bsnm{Zhu}, \binits{J.-Y.}},
\bauthor{\bsnm{Ermon}, \binits{S.}}:
\bctitle{{SDE}dit: Guided image synthesis and editing with stochastic
  differential equations}.
In: \bbtitle{International Conference on Learning Representations}
(\byear{2022})
\end{bchapter}
\endbibitem

\bibitem[\protect\citeauthoryear{Zhang et~al.}{2023}]{controlnet}
\begin{botherref}
\oauthor{\bsnm{Zhang}, \binits{L.}},
\oauthor{\bsnm{Rao}, \binits{A.}},
\oauthor{\bsnm{Agrawala}, \binits{M.}}:
Adding Conditional Control to Text-to-Image Diffusion Models
(2023)
\end{botherref}
\endbibitem

\bibitem[\protect\citeauthoryear{Saharia et~al.}{2021}]{sr3}
\begin{botherref}
\oauthor{\bsnm{Saharia}, \binits{C.}},
\oauthor{\bsnm{Ho}, \binits{J.}},
\oauthor{\bsnm{Chan}, \binits{W.}},
\oauthor{\bsnm{Salimans}, \binits{T.}},
\oauthor{\bsnm{Fleet}, \binits{D.J.}},
\oauthor{\bsnm{Norouzi}, \binits{M.}}:
Image super-resolution via iterative refinement.
arXiv:2104.07636
(2021)
\end{botherref}
\endbibitem

\bibitem[\protect\citeauthoryear{Saharia et~al.}{2022}]{palette}
\begin{bchapter}
\bauthor{\bsnm{Saharia}, \binits{C.}},
\bauthor{\bsnm{Chan}, \binits{W.}},
\bauthor{\bsnm{Chang}, \binits{H.}},
\bauthor{\bsnm{Lee}, \binits{C.}},
\bauthor{\bsnm{Ho}, \binits{J.}},
\bauthor{\bsnm{Salimans}, \binits{T.}},
\bauthor{\bsnm{Fleet}, \binits{D.}},
\bauthor{\bsnm{Norouzi}, \binits{M.}}:
\bctitle{Palette: Image-to-image diffusion models}.
In: \bbtitle{ACM SIGGRAPH 2022 Conference Proceedings}.
\bsertitle{SIGGRAPH '22}.
\bpublisher{Association for Computing Machinery},
\blocation{New York, NY, USA}
(\byear{2022}).
\doiurl{10.1145/3528233.3530757} .
\burl{https://doi.org/10.1145/3528233.3530757}
\end{bchapter}
\endbibitem

\bibitem[\protect\citeauthoryear{Yang et~al.}{2023}]{yang2023rerender}
\begin{bchapter}
\bauthor{\bsnm{Yang}, \binits{S.}},
\bauthor{\bsnm{Zhou}, \binits{Y.}},
\bauthor{\bsnm{Liu}, \binits{Z.}},
\bauthor{},
\bauthor{\bsnm{Loy}, \binits{C.C.}}:
\bctitle{Rerender a video: Zero-shot text-guided video-to-video translation}.
In: \bbtitle{ACM SIGGRAPH Asia Conference Proceedings}
(\byear{2023})
\end{bchapter}
\endbibitem

\bibitem[\protect\citeauthoryear{Jamriska}{2018}]{ebsynth}
\begin{botherref}
\oauthor{\bsnm{Jamriska}, \binits{O.}}:
Ebsynth: Fast Example-based Image Synthesis and Style Transfer.
GitHub
(2018)
\end{botherref}
\endbibitem

\bibitem[\protect\citeauthoryear{Liang et~al.}{2005}]{video2cartoon1}
\begin{bchapter}
\bauthor{\bsnm{Liang}, \binits{D.}},
\bauthor{\bsnm{Liu}, \binits{Y.}},
\bauthor{\bsnm{Huang}, \binits{Q.}},
\bauthor{\bsnm{Zhu}, \binits{G.}},
\bauthor{\bsnm{Jiang}, \binits{S.}},
\bauthor{\bsnm{Zhang}, \binits{Z.}},
\bauthor{\bsnm{Gao}, \binits{W.}}:
\bctitle{Video2cartoon: generating 3d cartoon from broadcast soccer video}.
In: \bbtitle{MULTIMEDIA '05}
(\byear{2005})
\end{bchapter}
\endbibitem

\bibitem[\protect\citeauthoryear{Ngo and Cai}{2008}]{video2cartoon2}
\begin{bchapter}
\bauthor{\bsnm{Ngo}, \binits{V.}},
\bauthor{\bsnm{Cai}, \binits{J.}}:
\bctitle{Converting 2d soccer video to 3d cartoon}.
In: \bbtitle{2008 10th International Conference on Control, Automation,
  Robotics and Vision},
pp. \bfpage{103}--\blpage{107}
(\byear{2008}).
\doiurl{10.1109/ICARCV.2008.4795500}
\end{bchapter}
\endbibitem

\bibitem[\protect\citeauthoryear{Weng et~al.}{2019}]{photowakeup}
\begin{bchapter}
\bauthor{\bsnm{Weng}, \binits{C.-Y.}},
\bauthor{\bsnm{Curless}, \binits{B.}},
\bauthor{\bsnm{Kemelmacher-Shlizerman}, \binits{I.}}:
\bctitle{Photo wake-up: 3d character animation from a single photo}.
In: \bbtitle{Proceedings of the IEEE/CVF Conference on Computer Vision and
  Pattern Recognition (CVPR)}
(\byear{2019})
\end{bchapter}
\endbibitem

\bibitem[\protect\citeauthoryear{Tous}{2023}]{pictonaut}
\begin{botherref}
\oauthor{\bsnm{Tous}, \binits{R.}}:
Pictonaut: movie cartoonization using 3d human pose estimation and gans.
Multimedia Tools and Applications,
1--15
(2023)
\doiurl{10.1007/s11042-023-14556-1}
\end{botherref}
\endbibitem

\bibitem[\protect\citeauthoryear{Ruben~Tous and Igual}{2023}]{completepose}
\begin{barticle}
\bauthor{\bsnm{Ruben~Tous}, \binits{J.N.}},
\bauthor{\bsnm{Igual}, \binits{L.}}:
\batitle{Human pose completion in partial body camera shots}.
\bjtitle{Journal of Experimental \& Theoretical Artificial Intelligence}
\bvolume{0}(\bissue{0}),
\bfpage{1}--\blpage{11}
(\byear{2023})
\doiurl{10.1080/0952813X.2023.2241575}
\end{barticle}
\endbibitem

\bibitem[\protect\citeauthoryear{Fi\v{s}er et~al.}{2017}]{Fiser17-SIG}
\begin{botherref}
\oauthor{\bsnm{Fi\v{s}er}, \binits{J.}},
\oauthor{\bsnm{Jamri\v{s}ka}, \binits{O.}},
\oauthor{\bsnm{Simons}, \binits{D.}},
\oauthor{\bsnm{Shechtman}, \binits{E.}},
\oauthor{\bsnm{Lu}, \binits{J.}},
\oauthor{\bsnm{Asente}, \binits{P.}},
\oauthor{\bsnm{Luk\'{a}\v{c}}, \binits{M.}},
\oauthor{\bsnm{S\'{y}kora}, \binits{D.}}:
Example-based synthesis of stylized facial animations.
ACM Transactions on Graphics
\textbf{36}(4)
(2017)
\end{botherref}
\endbibitem

\bibitem[\protect\citeauthoryear{Gatys et~al.}{2015}]{styletransfer}
\begin{botherref}
\oauthor{\bsnm{Gatys}, \binits{L.A.}},
\oauthor{\bsnm{Ecker}, \binits{A.S.}},
\oauthor{\bsnm{Bethge}, \binits{M.}}:
A neural algorithm of artistic style.
CoRR
\textbf{abs/1508.06576}
(2015)
{\href{https://arxiv.org/abs/1508.06576}{{1508.06576}}}
\end{botherref}
\endbibitem

\bibitem[\protect\citeauthoryear{He et~al.}{2022}]{MAE}
\begin{bchapter}
\bauthor{\bsnm{He}, \binits{K.}},
\bauthor{\bsnm{Chen}, \binits{X.}},
\bauthor{\bsnm{Xie}, \binits{S.}},
\bauthor{\bsnm{Li}, \binits{Y.}},
\bauthor{\bsnm{Dollar}, \binits{P.}},
\bauthor{\bsnm{Girshick}, \binits{R.}}:
\bctitle{Masked autoencoders are scalable vision learners}.
In: \bbtitle{2022 IEEE/CVF Conference on Computer Vision and Pattern
  Recognition (CVPR)},
pp. \bfpage{15979}--\blpage{15988}
(\byear{2022}).
\doiurl{10.1109/CVPR52688.2022.01553}
\end{bchapter}
\endbibitem

\bibitem[\protect\citeauthoryear{Dosovitskiy et~al.}{2021}]{ViT}
\begin{bchapter}
\bauthor{\bsnm{Dosovitskiy}, \binits{A.}},
\bauthor{\bsnm{Beyer}, \binits{L.}},
\bauthor{\bsnm{Kolesnikov}, \binits{A.}},
\bauthor{\bsnm{Weissenborn}, \binits{D.}},
\bauthor{\bsnm{Zhai}, \binits{X.}},
\bauthor{\bsnm{Unterthiner}, \binits{T.}},
\bauthor{\bsnm{Dehghani}, \binits{M.}},
\bauthor{\bsnm{Minderer}, \binits{M.}},
\bauthor{\bsnm{Heigold}, \binits{G.}},
\bauthor{\bsnm{Gelly}, \binits{S.}},
\bauthor{\bsnm{Uszkoreit}, \binits{J.}},
\bauthor{\bsnm{Houlsby}, \binits{N.}}:
\bctitle{An image is worth 16x16 words: Transformers for image recognition at
  scale}.
In: \bbtitle{9th International Conference on Learning Representations, {ICLR}
  2021, Virtual Event, Austria, May 3-7, 2021}.
\bpublisher{OpenReview.net}, \blocation{???}
(\byear{2021})
\end{bchapter}
\endbibitem

\bibitem[\protect\citeauthoryear{Radford et~al.}{2021}]{CLIP}
\begin{bchapter}
\bauthor{\bsnm{Radford}, \binits{A.}},
\bauthor{\bsnm{Kim}, \binits{J.W.}},
\bauthor{\bsnm{Hallacy}, \binits{C.}},
\bauthor{\bsnm{Ramesh}, \binits{A.}},
\bauthor{\bsnm{Goh}, \binits{G.}},
\bauthor{\bsnm{Agarwal}, \binits{S.}},
\bauthor{\bsnm{Sastry}, \binits{G.}},
\bauthor{\bsnm{Askell}, \binits{A.}},
\bauthor{\bsnm{Mishkin}, \binits{P.}},
\bauthor{\bsnm{Clark}, \binits{J.}},
\bauthor{\bsnm{Krueger}, \binits{G.}},
\bauthor{\bsnm{Sutskever}, \binits{I.}}:
\bctitle{Learning transferable visual models from natural language
  supervision}.
In: \beditor{\bsnm{Meila}, \binits{M.}},
\beditor{\bsnm{Zhang}, \binits{T.}} (eds.)
\bbtitle{Proceedings of the 38th International Conference on Machine Learning}.
\bsertitle{Proceedings of Machine Learning Research},
vol. \bseriesno{139},
pp. \bfpage{8748}--\blpage{8763}.
\bpublisher{PMLR}, \blocation{???}
(\byear{2021})
\end{bchapter}
\endbibitem

\bibitem[\protect\citeauthoryear{Carion
  et~al.}{2020}]{segmentation_transformers}
\begin{bchapter}
\bauthor{\bsnm{Carion}, \binits{N.}},
\bauthor{\bsnm{Massa}, \binits{F.}},
\bauthor{\bsnm{Synnaeve}, \binits{G.}},
\bauthor{\bsnm{Usunier}, \binits{N.}},
\bauthor{\bsnm{Kirillov}, \binits{A.}},
\bauthor{\bsnm{Zagoruyko}, \binits{S.}}:
\bctitle{End-to-end object detection with transformers}.
In: \bbtitle{Computer Vision – ECCV 2020: 16th European Conference, Glasgow,
  UK, August 23–28, 2020, Proceedings, Part I},
pp. \bfpage{213}--\blpage{229}.
\bpublisher{Springer},
\blocation{Berlin, Heidelberg}
(\byear{2020}).
\doiurl{10.1007/978-3-030-58452-8\_13} .
\burl{https://doi.org/10.1007/978-3-030-58452-8\_13}
\end{bchapter}
\endbibitem

\bibitem[\protect\citeauthoryear{Yang and Yang}{2022}]{DeAOT}
\begin{bchapter}
\bauthor{\bsnm{Yang}, \binits{Z.}},
\bauthor{\bsnm{Yang}, \binits{Y.}}:
\bctitle{Decoupling features in hierarchical propagation for video object
  segmentation}.
In: \bbtitle{Advances in Neural Information Processing Systems (NeurIPS)}
(\byear{2022})
\end{bchapter}
\endbibitem

\bibitem[\protect\citeauthoryear{Liu et~al.}{2023}]{Grounding-DINO}
\begin{botherref}
\oauthor{\bsnm{Liu}, \binits{S.}},
\oauthor{\bsnm{Zeng}, \binits{Z.}},
\oauthor{\bsnm{Ren}, \binits{T.}},
\oauthor{\bsnm{Li}, \binits{F.}},
\oauthor{\bsnm{Zhang}, \binits{H.}},
\oauthor{\bsnm{Yang}, \binits{J.}},
\oauthor{\bsnm{Li}, \binits{C.}},
\oauthor{\bsnm{Yang}, \binits{J.}},
\oauthor{\bsnm{Su}, \binits{H.}},
\oauthor{\bsnm{Zhu}, \binits{J.}}, et al.:
Grounding dino: Marrying dino with grounded pre-training for open-set object
  detection.
arXiv preprint arXiv:2303.05499
(2023)
\end{botherref}
\endbibitem

\bibitem[\protect\citeauthoryear{Suzuki and be}{1985}]{findcontours}
\begin{barticle}
\bauthor{\bsnm{Suzuki}, \binits{S.}},
\bauthor{\bsnm{be}, \binits{K.}}:
\batitle{Topological structural analysis of digitized binary images by border
  following}.
\bjtitle{Computer Vision, Graphics, and Image Processing}
\bvolume{30}(\bissue{1}),
\bfpage{32}--\blpage{46}
(\byear{1985})
\doiurl{10.1016/0734-189X(85)90016-7}
\end{barticle}
\endbibitem

\bibitem[\protect\citeauthoryear{Kazemi and Sullivan}{2014}]{dlib_HOG2}
\begin{bchapter}
\bauthor{\bsnm{Kazemi}, \binits{V.}},
\bauthor{\bsnm{Sullivan}, \binits{J.}}:
\bctitle{One millisecond face alignment with an ensemble of regression trees}.
In: \bbtitle{2014 IEEE Conference on Computer Vision and Pattern Recognition},
pp. \bfpage{1867}--\blpage{1874}
(\byear{2014}).
\doiurl{10.1109/CVPR.2014.241}
\end{bchapter}
\endbibitem

\bibitem[\protect\citeauthoryear{Dalal and Triggs}{2005}]{HOG}
\begin{bchapter}
\bauthor{\bsnm{Dalal}, \binits{N.}},
\bauthor{\bsnm{Triggs}, \binits{B.}}:
\bctitle{Histograms of oriented gradients for human detection}.
In: \bbtitle{2005 IEEE Computer Society Conference on Computer Vision and
  Pattern Recognition (CVPR'05)},
vol. \bseriesno{1},
pp. \bfpage{886}--\blpage{8931}
(\byear{2005}).
\doiurl{10.1109/CVPR.2005.177}
\end{bchapter}
\endbibitem

\bibitem[\protect\citeauthoryear{Soomro et~al.}{2012}]{ucf101-dataset}
\begin{botherref}
\oauthor{\bsnm{Soomro}, \binits{K.}},
\oauthor{\bsnm{Zamir}, \binits{A.R.}},
\oauthor{\bsnm{Shah}, \binits{M.}}:
{UCF101:} {A} dataset of 101 human actions classes from videos in the wild.
CoRR
\textbf{abs/1212.0402}
(2012)
{\href{https://arxiv.org/abs/1212.0402}{{1212.0402}}}
\end{botherref}
\endbibitem

\bibitem[\protect\citeauthoryear{Zablotskaia
  et~al.}{2019}]{fashion-video-dataset}
\begin{bchapter}
\bauthor{\bsnm{Zablotskaia}, \binits{P.}},
\bauthor{\bsnm{Siarohin}, \binits{A.}},
\bauthor{\bsnm{Zhao}, \binits{B.}},
\bauthor{\bsnm{Sigal}, \binits{L.}}:
\bctitle{Dwnet: Dense warp-based network for pose-guided human video
  generation}.
In: \bbtitle{30th British Machine Vision Conference 2019, {BMVC} 2019, Cardiff,
  UK, September 9-12, 2019},
p. \bfpage{51}.
\bpublisher{{BMVA} Press}, \blocation{???}
(\byear{2019})
\end{bchapter}
\endbibitem

\bibitem[\protect\citeauthoryear{Kay et~al.}{2017}]{kinetics-dataset}
\begin{botherref}
\oauthor{\bsnm{Kay}, \binits{W.}},
\oauthor{\bsnm{Carreira}, \binits{J.}},
\oauthor{\bsnm{Simonyan}, \binits{K.}},
\oauthor{\bsnm{Zhang}, \binits{B.}},
\oauthor{\bsnm{Hillier}, \binits{C.}},
\oauthor{\bsnm{Vijayanarasimhan}, \binits{S.}},
\oauthor{\bsnm{Viola}, \binits{F.}},
\oauthor{\bsnm{Green}, \binits{T.}},
\oauthor{\bsnm{Back}, \binits{T.}},
\oauthor{\bsnm{Natsev}, \binits{P.}},
\oauthor{\bsnm{Suleyman}, \binits{M.}},
\oauthor{\bsnm{Zisserman}, \binits{A.}}:
The kinetics human action video dataset.
CoRR
\textbf{abs/1705.06950}
(2017)
{\href{https://arxiv.org/abs/1705.06950}{{1705.06950}}}
\end{botherref}
\endbibitem

\bibitem[\protect\citeauthoryear{Tous}{}]{lester}
\begin{botherref}
\oauthor{\bsnm{Tous}, \binits{R.}}:
Lester Project Home Page.
Accessed: 2023-12-20.
\url{https://github.com/rtous/lester}
\end{botherref}
\endbibitem

\bibitem[\protect\citeauthoryear{ITU}{2016}]{p800}
\begin{botherref}
\oauthor{\bsnm{ITU}}:
Mean opinion score (MOS) terminology. ITU-T Recommendation P.800.1 Methods for
  objective and subjective assessment of speech and video quality
(2016).
\url{https://www.itu.int/rec/dologin\_pub.asp?lang=s\&id=T-REC-P.800.1-201607-I!!PDF-E\&type=items}
\end{botherref}
\endbibitem

\bibitem[\protect\citeauthoryear{ITU}{2008}]{p910}
\begin{botherref}
\oauthor{\bsnm{ITU}}:
Subjective Video Quality Assessment Methods for Multimedia Applications. ITU-T
  Recommendation P.910
(2008).
\url{https://www.itu.int/rec/T-REC-P.910}
\end{botherref}
\endbibitem

\end{thebibliography}


\end{document}